\documentclass[11pt]{article}
\usepackage[a4paper, margin=2.5cm]{geometry}
\usepackage[T1]{fontenc}
\usepackage[normalem]{ulem}

\usepackage[ruled,vlined]{algorithm2e}
\usepackage{graphicx}
\usepackage{booktabs}
\usepackage{xcolor}
\usepackage{rotating}

\usepackage[dvipsnames,svgnames,x11names]{xcolor}

\colorlet{RED}{red}
\colorlet{WHITE}{white}
\colorlet{FORESTGREEN}{ForestGreen}
\colorlet{MAROON}{Maroon}

\usepackage[nocompress]{cite}
\usepackage{balance}
\usepackage{listings}

\lstdefinelanguage{XML}
{
basicstyle=\ttfamily\footnotesize,
  morestring=[b]",
  moredelim=[s][\bfseries\color{Maroon}]{<}{\ },
  moredelim=[s][\bfseries\color{Maroon}]{</}{>},
  moredelim=[l][\bfseries\color{Maroon}]{/>},
  moredelim=[l][\bfseries\color{Maroon}]{>},
  morecomment=[s]{<?}{?>},
  morecomment=[s]{<!--}{-->},
  commentstyle=\color{gray},
  stringstyle=\color{blue},
  identifierstyle=\color{red}
}

\usepackage{moreverb}
\usepackage[nounderscore]{syntax}
\graphicspath{{./figures/}}
\DeclareGraphicsExtensions{.pdf}
\usepackage[cmex10]{amsmath}
\usepackage{amssymb}
\usepackage{mathtools}
\usepackage{amsthm}
\usepackage{amsfonts}

\usepackage{subfig}

\usepackage{multirow}
\usepackage{rotating}
\usepackage{booktabs}
\usepackage{colortbl}
\usepackage{tablefootnote}

\usepackage{array}
\newcolumntype{L}[1]{>{\raggedright\let\newline\\\arraybackslash\hspace{0pt}}m{#1}}
\newcolumntype{C}[1]{>{\centering\let\newline\\\arraybackslash\hspace{0pt}}m{#1}}
\newcolumntype{R}[1]{>{\raggedleft\let\newline\\\arraybackslash\hspace{0pt}}m{#1}}

\usepackage[pdftex,colorlinks=true,urlcolor=blue,citecolor=blue]{hyperref}
\usepackage{xspace}

\usepackage{enumitem}

\usepackage[utf8]{inputenc}
\usepackage[T1]{fontenc}
\usepackage{url}
\usepackage{booktabs}
\usepackage{amsfonts}
\usepackage{nicefrac}
\usepackage{lmodern}
\usepackage{microtype}
\usepackage{lipsum}
\usepackage{rotating}
\usepackage{multirow}
\graphicspath{ {./images/} }

\newcommand{\uvhw}{UVH-26\xspace}
\newcommand{\uvh}{UVH\xspace}
\newcommand{\UVH}{Urban Vision Hackathon\xspace}
\newcommand{\uvhmv}{\uvhw-MV\xspace}
\newcommand{\uvhst}{\uvhw-ST\xspace}

\begin{document}

\title{\textbf{The Urban Vision Hackathon Dataset and Models:}\\
Towards Image Annotations and Accurate Vision Models
for Indian Traffic\\
\textsc{\large\textit{Preliminary Dataset Release, \uvhw-v1.0}}}

\author{
    Akash Sharma$^{1}$, Chinmay Mhatre$^{2}$, Sankalp Gawali$^{1}$,
    Ruthvik Bokkasam$^{1}$,\\ Brij Kishore$^{4}$, Vishwajeet Pattanaik$^{2}$, 
    Tarun Rambha$^{2}$, Abdul R. Pinjari$^{2}$,\\ Vijay Kovvali$^{2}$,
    Anirban Chakraborty$^{1}$, Punit Rathore$^{3,2}$,\\
    Raghu Krishnapuram$^{4,2}$ and Yogesh Simmhan$^{1}$\\~\\
    \small\em AI for Integrated Mobility (AIM) Team\\
    \small\em $^{1}$Department of Computation and Data Sciences (CDS)\\
    \small\em $^{2}$Centre for Infrastructure, Sustainable Transportation and Urban Planning (CiSTUP)\\
    \small\em $^{3}$Robert Bosch Centre for Cyberphysical Systems (RBCCPS)\\
    \small\em $^{4}$Centre for Data for Public Good (CDPG)\\
    \em Indian Institute of Science, Bengaluru, India\\~\\
    \texttt{Email: \{akashsharma, simmhan\}@iisc.ac.in}
}
\date{4 Nov, 2025}

\maketitle

\begin{abstract}
    This report describes the \textit{\uvhw dataset}, the first public release by AIM@IISc of a large-scale dataset of annotated traffic-camera images from India. The dataset comprises $26,646$ high-resolution (1080p) images sampled from $\approx 2800$ Bengaluru's Safe-City CCTV cameras over a 4-week period, and subsequently annotated through a crowdsourced hackathon involving 565 college students from across India. In total, $1.8$ million bounding boxes were labeled across $14$ vehicle classes specific to India: Cycle, 2-Wheeler (Motorcycle), 3-Wheeler (Auto-rickshaw), LCV (Light Commercial Vehicles), Van, Tempo-traveller, Hatchback, Sedan, SUV, MUV, Mini-bus, Bus, Truck, and Other.
    Of these, $\approx283k$--$316k$ consensus ground truth bounding boxes and labels were derived for distinct objects in the $26k$ images using Majority Voting and STAPLE algorithms.
    Further, we train multiple contemporary detectors, including
    YOLO11-S/X,
    RT-DETR-S/X,
    and DAMO-YOLO-T/L using these datasets,
    and report accuracy based on mAP50, mAP75 and mAP50:95. Models trained on \uvhw achieve $\approx8.4$--$31.5\%$ improvements in mAP50:95 over equivalent baseline models trained on COCO dataset, with RT-DETR-X showing the best performance at $0.67$ (mAP50:95) as compared to $0.40$ for COCO-trained weights for common classes (Car, Bus, and Truck). This demonstrates the benefits of domain-specific training data for Indian traffic scenarios. The release package provides the $26k$ images with consensus annotations based on Majority Voting (\uvhmv) and STAPLE (\uvhst), and the $6$ fine-tuned YOLO and DETR models on each of these datasets.
    By capturing the heterogeneity of Indian urban mobility directly from operational traffic-camera streams, \uvhw addresses a critical gap in existing global benchmarks, and offers a foundation for advancing detection, classification, and deployment of intelligent transportation systems in emerging nations with complex traffic conditions.
\end{abstract}

\section{Introduction}

Intelligent Transportation Systems (ITS) increasingly depend on robust vehicle detection and classification models to enable traffic monitoring, policy enforcement, and urban planning~\cite{cheng2021survey,li2021surveyits}. The performance of these models is critically influenced by the quality and relevance of the training data. While large-scale object detection datasets such as COCO~\cite{lin2014coco} and Objects365~\cite{shao2019objects365} have significantly advanced general-purpose object detection, their applicability to traffic scenarios in developing regions such as India remains limited. These existing datasets predominantly feature urban environments from developed or western countries, whose organized traffic conditions differ markedly from the heterogeneous, high-density and complex traffic conditions observed in mega-cities of South Asia and developing nations.

Urban traffic in countries like India presents unique challenges, including extreme vehicle density, non-standard driving behavior (e.g., failure to follow lane restrictions), and a diverse mix of vehicle types including auto-rickshaws/tuk-tuks (3-Wheelers), motorcycles/scooters (2-Wheelers), and Light Commercial Vehicles (LCVs). Popular and state-of-the-art (SOTA) object detection models such as YOLO~\cite{jocher2023yolov8} and RT-DETR~\cite{zhao2024rtdetr} are trained on a wider class of generic image datasets such as COCO~\cite{lin2014coco} and Objects365~\cite{shao2019objects365}, and whose traffic and vehicle related images tend to be from a different subset of vehicle types and from more organized traffic flow.
These are less effective when used directly for vehicle detection and classification in complex traffic environments like India. This highlights the need for region-specific datasets that capture the diversity of local traffic conditions~\cite{varma2019idd, agarwal2020bdd100k,wu2022densetraf,raj2021indian,deng2024trafficcam}.

To address this gap, the \textit{AI for Integrated Mobility (AIM)} team at the Indian Institute of Science (IISc) is releasing \textbf{\uvhw}, a new public dataset of $26,678$ annotated 1080p high-resolution traffic images from India, curated from CCTV footage collected in collaboration with the Bengaluru Traffic Police. The 
images represent complex urban traffic scenes in Bengaluru, a mega city with over 10 million residents and, by some measures, with the third slowest traffic in the world~\cite{tomtom2024trafficindex_bengaluru}.
These images are annotated with $283k$ and $316k$ bounding boxes/labels using two alternate consensus algorithms,
by using 14 fine-grained India-specific vehicle classes, broadly based on the Indian Road Congress classification (Table~\ref{tab:vehicle_category_description}).
We blur the faces in the images to respect privacy.

Given the high cost and effort associated with expert annotations, we developed a gamified web-based platform to crowdsource the annotation process. Over 550 student volunteers from across India actively participated in this \textit{Urban Vision Hackathon (UVH)}~\footnote{https://airawat-mobility.github.io/hack/} held in May and June, 2025, 
incentivized through competitive scoring, leaderboards, daily/weekly prizes, and internships. To ease the annotation overhead and maintain consistency, we adopted a model-assisted labeling approach using a pre-trained RT-DETRv2-X~\cite{zhao2024rtdetr} detector using 
$\approx 3000$ expert-labeled images
to generate pre-annotations. The participants could then validate, correct, or supplement these bounding box and vehicle class predictions, significantly reducing manual effort while having a human-in-the-loop validation.

To increase the quality of crowdsourced volunteer-driven annotations, the same image is shown to multiple participants. Further, we occasionally embed ``gold'' images with known ground-truth annotations, which are visually indistinguishable from the other images, but are used to estimate the running accuracy of the participants.
These accuracy metrics are used to estimate a reliable consensus ground truth using a simple \textit{majority voting} and the more complex Expectation Maximization (EM) based 
\textit{STAPLE} algorithm~\cite{warfield2004simultaneous}.

Lastly, to help bootstrap the AI benefits from this dataset, we also release \textit{$6$ detection models} that are pre-trained using this dataset, based on contemporary model architectures and with diverse footprints, to allow deployment on heterogeneous accelerated edge and server platforms. These 
models are based on  YOLOv11-S and -X~\cite{jocher2023yolov8}, DAMO-YOLO-T and -L~\cite{xu2022damo}, RT-DETRv2-S, and -X~\cite{carion2020detr}. We report their accuracy based on mAP50, mAP75, and mAP50:95. These models trained on \uvhw achieve $\approx 8.4$--$31.5\%$ improvement on mAP50:95 over equivalent baseline models pre-trained only using the COCO dataset, with the best performing model, RT-DETRv2, showing a $27\%$ improvement.

The primary goal of releasing this large-scale \uvhw dataset and associated models is to help the community design better computer vision models for vehicle detection in Indian traffic conditions, and complement other such datasets that are emerging~\cite{varma2019idd, deng2024trafficcam}. These can then serve as a building block for more advanced AI-driven analytics for intelligent traffic management, to help reduce congestion, improve road safety, and enhance sustainability in India and other developing countries in the global south.

In the rest of this report, we detail the dataset creation and annotation methodology (\S~\ref{sec:dataset}), the consensus algorithms used to generate the final annotations from the crowdsourced ones (\S~\ref{sec:consensus}), the methodology used for training the detection and classification models using these images (\S~\ref{sec:train}), and lastly performance metrics comparing the results from these \uvhw trained models against those from contemporary detection architectures (\S~\ref{sec:benchmarks}). Additional details are provided in the Appendices.

The dataset are made publicly available at \url{https://huggingface.co/datasets/iisc-aim/UVH-26} (branch \texttt{v1.0}) under a Creative Commons Attribution 4.0 International License, and while the models are made publicly available \url{https://huggingface.co/iisc-aim/UVH-26} (branch \texttt{v1.0}) under a
Apache 2.0 License. More details are provided in Appendix~\ref{app:release}.

\section{Dataset Design and Crowdsourcing Workflow \label{sec:dataset}}
In this section, we describe the Version 1.0 release of the \textit{\UVH 26k dataset (\uvhw)}, which consists of $26,646$ frames that have been annotated during the first week of a 
five-week nationwide annotation challenge.
\subsection{Base Images from CCTV Cameras}
The base images in this release are sourced from $\approx 2800$ cameras installed by the Bengaluru Police under the `Safe City' urban safety project, and repurposed here for traffic analytics. The cameras are spread across the city, with higher density along peripheral traffic corridors and in the central business district. They include both road junctions (intersections) and mid-block viewpoints. The camera feeds use wide fields of view to support urban safety needs and are therefore not aligned to precisely monitor traffic lanes. As a result, the scenes exhibit greater perspective variation and occlusions, making detection more challenging.
Figure~\ref{fig:camera-map} in Appendix~\ref{app:cameras} shows the a map of the cameras used for images present in the \uvhw data.

Frames were sampled between 06:00 and 18:00 IST of a 25-day period in February, 2025, covering daytime urban traffic when cameras consistently capture color images and the traffic conditions are most active. Night-time frames, which are typically monochrome due to limitations of CCTV sensors, were excluded. 
Each frame is stored as a $1920\times1080$ RGB image, with a tiny fraction at a lower resolution. 

\begin{figure}[t]
    \centering
        \includegraphics[width=0.45\textwidth]{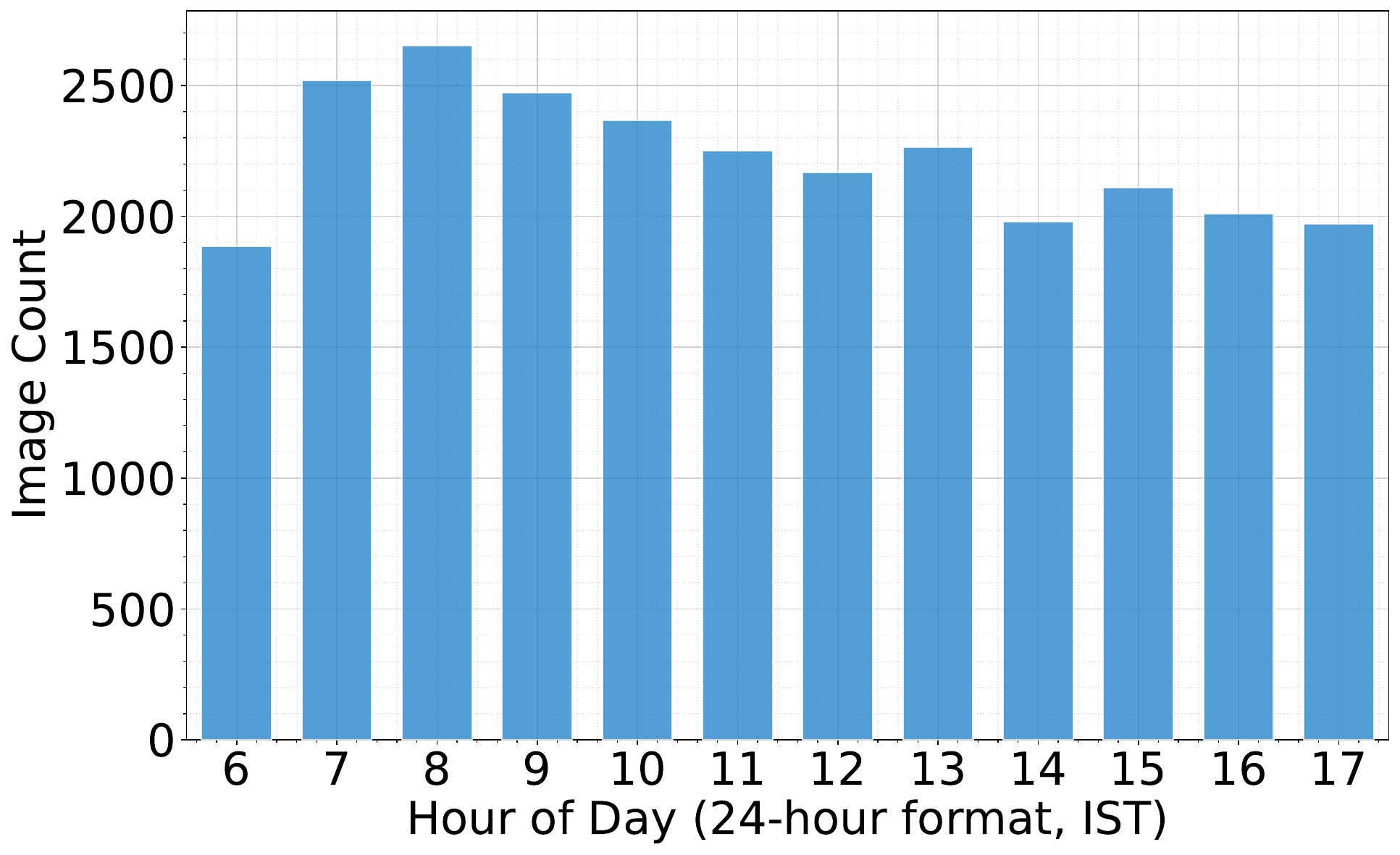}
    \caption{Time of day distribution of images in \uvhw across 25 days.}
    \label{fig:hour_and_date_distributions}
\end{figure}

From this pool of frames, we further select $100k$ images with complex scenes and providing divergent detections by the baseline pre-annotation models, as we describe later, to ensure that manual annotations are done only for challenging visual scenarios. 
Figure~\ref{fig:hour_and_date_distributions} shows the distribution of images in \uvhw across different hours of the day, spanning 25 days.

\begin{figure}[t]
    \centering
    
    \includegraphics[width=0.9\textwidth]{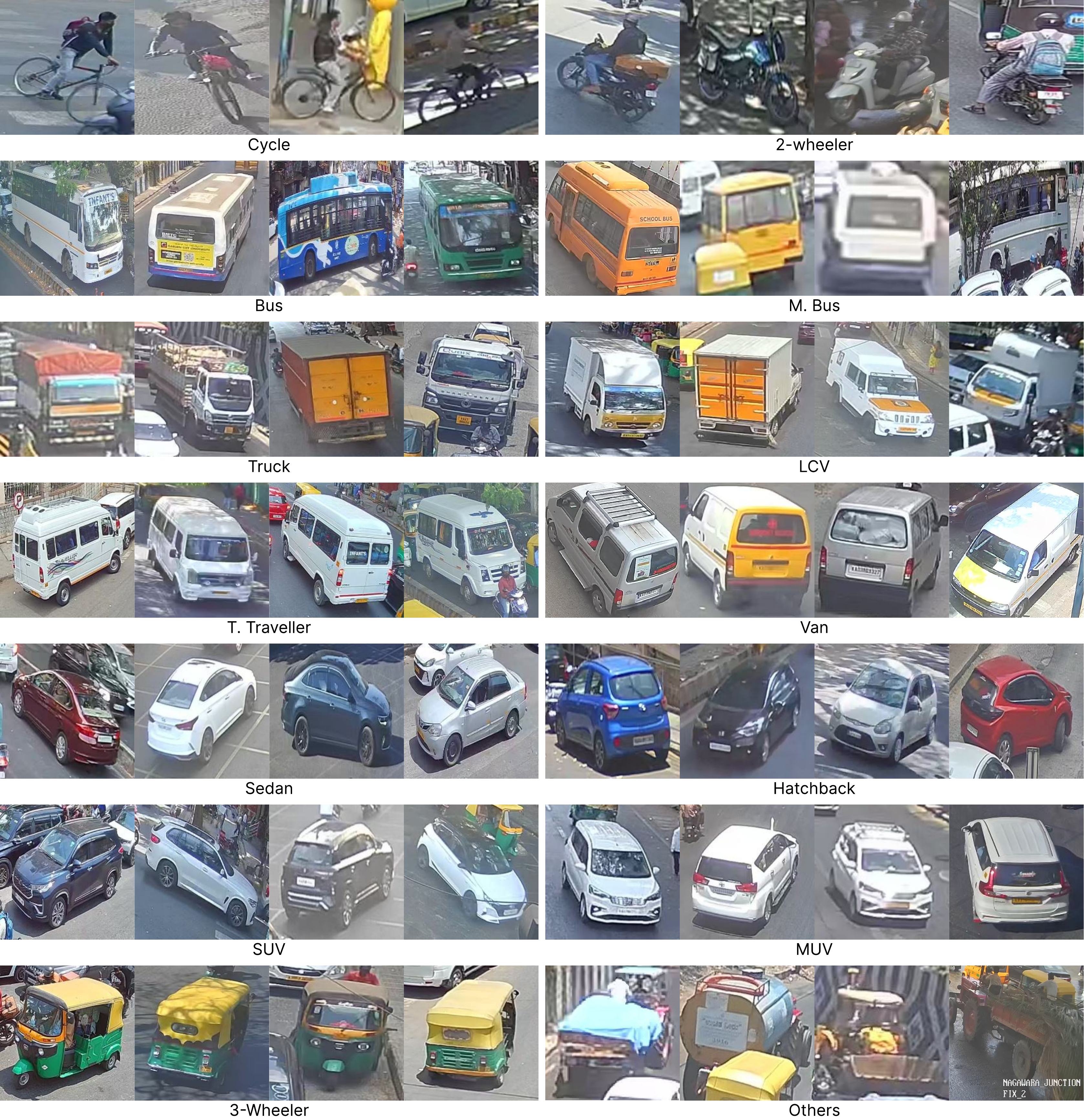}
    \caption{Example cropped images of each of 14 classes in the \uvhw dataset.}
    \label{fig:example_classes}
\end{figure}

\subsection{Vehicle Classes}

We focus on $14$ fine-grained vehicle classes that reflect the diversity of India's vehicle fleet, as defined by the Indian Road Congress~\cite{IRC}: \emph{Hatchback}, \emph{Sedan}, \emph{SUV}, \emph{MUV}, \emph{Bus}, \emph{Truck}, \emph{3-Wheeler}, \emph{2-Wheeler}, \emph{LCV}, \emph{M. Bus} (Mini-bus), \emph{T. Traveller} (Tempo-traveller), \emph{Cycle} (Bicycle), \emph{Van}, and \emph{Other}. Vehicles that could not be cleanly mapped to any of the standard classes were marked as ``Other''. We will refer to these vehicle classes as the \textit{\uvh classes}. Detailed descriptions of each vehicle class is provided in Table~\ref{tab:vehicle_category_description} of the Appendix, and cropped samples of the vehicles in each class are shown in Figure~\ref{fig:example_classes}.

\subsection{Expert Annotated Gold Dataset}

We first curate a ``Gold Dataset'' of $\approx3,000$ images sampled from $\approx 200$ cameras
and are manually annotated by paid experts using the \uvhw vehicle classes. This serves two purposes:
\begin{enumerate}
    \item \textit{Fine-tuning models for pre-annotations:} We use the expert-labeled data to fine-tune SOTA object detection models that are used to generate reliable pre-annotations for dataset that are subsequently annotated through crowdsourcing. This reduces the annotation burden on the participants.
    \item \textit{Quality control:} A subset of these expert-annotated images are embedded within the crowdsourcing workflow after randomly flipping labels to evaluate the participant's accuracy, and maintain high-quality human-in-the-loop validation.
\end{enumerate}

This gold data is not being made available publicly.

\subsection{Fine-tuned Models}

We fine-tuned several popular and SOTA object detection models using the gold dataset for the 14 classes of interest:

YOLOv8-X/N~\cite{jocher2023yolov8}, YOLOv11-X/N~\cite{jocher2024yolov11}, RT-DETR-X~\cite{zhao2024rtdetr}, D-FINE-X~\cite{peng2024dfine}, and DAMO-YOLO-X~\cite{xu2022damo}. Each model was initialized with weights pre-trained on COCO and subsequently fine-tuned using our gold dataset. About 2700 images from the gold dataset were used for training, while approximately 300 images were retained for testing.
This fine-tuning was necessary since the base models trained on the COCO dataset performed poorly on Indian traffic images.
Further, these models did not recognize certain vehicle classes unique to India and present in our \uvh classes, such as 3-wheelers and LCVs.
Among these, the best-performing fine-tuned model, RT-DETR-X, achieved a mean average precision (mAP@0.50:0.95) of $\approx 0.70$.

The predictions from all fine-tuned models were used to select a subset of images from the base $100k$ collection for crowdsourced annotation, based on disagreement and difficulty, as described next. Further, the best-performing RT-DETR-X model was used to pre-annotate images shown to the crowdsourcing participants.

\subsection{Disagreement and Image Difficulty}
\label{sec: disagreement_difficulty}

Crowdsourced annotations are limited to only images with complex scenes and those that are challenging for the fine-tuned models.
We use two complementary notions: image disagreement and image difficulty. \textit{Image disagreement} is the extent of prediction variation across multiple fine-tuned models for the same image, and provides a measure of ambiguity or confusion in the data for even SOTA models; and \textit{Image difficulty} serves as a proxy for factors such as occlusion, number of vehicles present, and small sized bounding boxes, 
ensuring that the dataset captures a diverse range of challenging scenarios.

\paragraph{Disagreement score.}
The \textit{disagreement score ($D_i$)} for an image captures two aspects of inter-model variability, where models disagree on the 
counts for each class \textit{(Per-class count disagreement $N_{dci}$)}, 
and the class with the largest pairwise disagreement \textit{(Maximum pairwise class-count disagreements $M_{mdi}$)}. 
This balances the aggregate count variance with the worst-case per-class pairwise disagreement.
These are described in Appendix~\ref{app:difficulty}.

\paragraph{Composite difficulty score.} While disagreement measures \emph{inter-model uncertainty}, it does not by itself quantify visual complexity. To ensure annotator workload was balanced 
we computed an image \emph{difficulty score ($\Delta_i$)} that captures intrinsic visual factors (object count, density, overlap) together with model disagreement.
We use several factors to compute the difficulty score for an image: the normalized count of bounding-boxes in the image ($\bar{C}$) -- more boxes means a harder image; the bounding-box sizes in the image ($M_{bbox\_size}$) -- smaller boxes make it harder to label; the density of bounding-boxes in the image ($M_{bbox\_density}$) -- higher density is more difficult; and the IoU overlap between bounding boxes $M_{iou\_overlap}$ -- more overlap indicates more occlusion and visual complexity.  
These are described in Appendix~\ref{app:difficulty}.

\subsection{Crowdsourced Annotations}\label{sec:cowdsourcing}

\begin{figure}[t]
    \centering
    \includegraphics[width=0.8\textwidth]{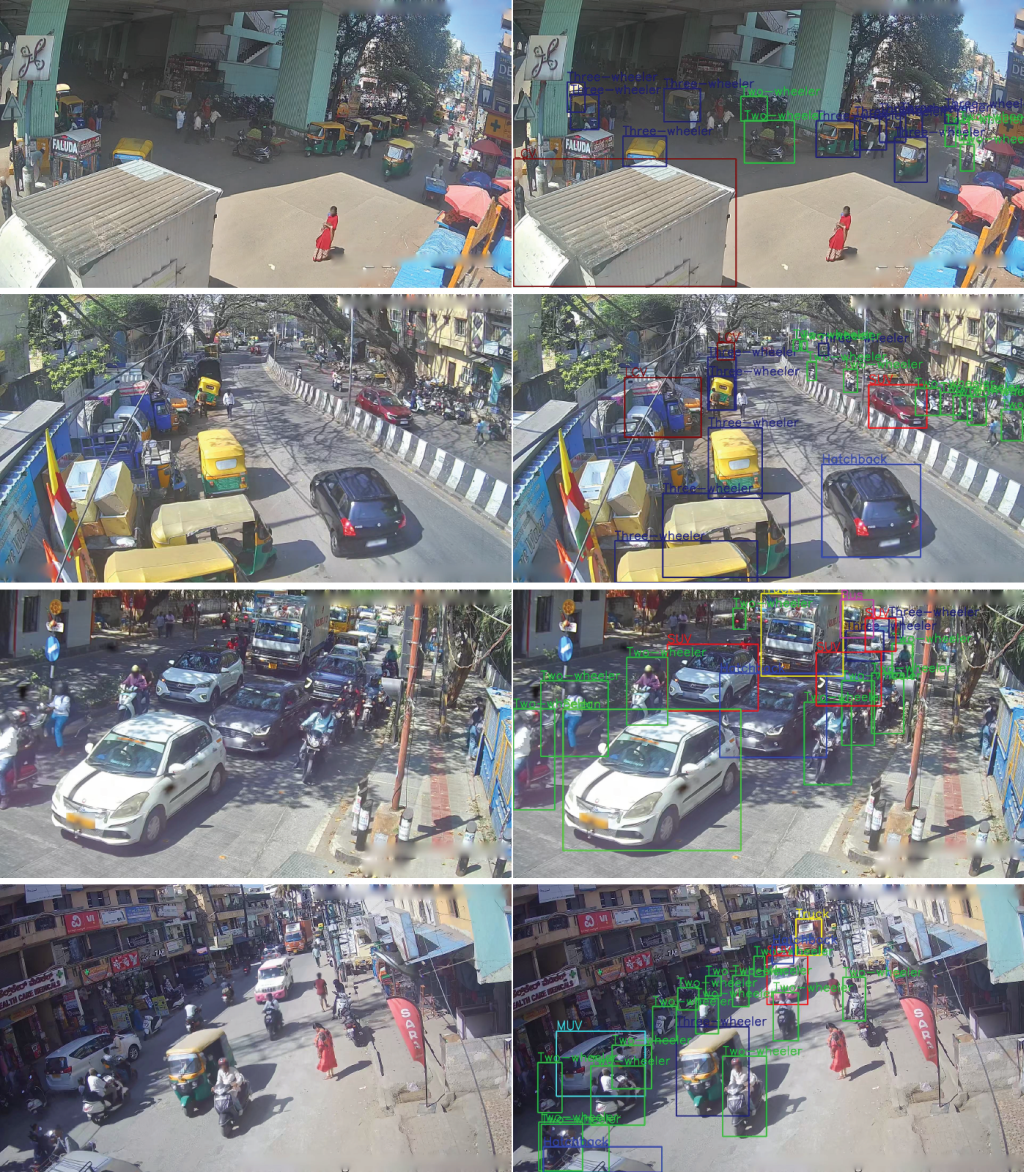}
    \caption{Example base images (left) and their pre-annotations with bounding box and label (right).}
    \label{fig:example_classes_2}
\end{figure}

Obtaining large-scale expert-annotated images is prohibitively expensive and time-consuming. To address this challenge, we used a crowdsourcing approach, which enables large-scale annotation by distributing tasks across many annotators while maintaining quality through careful task design, validation and consensus. Specifically, we hosted the \UVH (\uvh), an interactive and gamified annotation challenge,
with participants progressing through a map of Bengaluru and annotating traffic images
using a custom browser-based web interface.
About 568 participants from across India, mostly under-graduate students in teams of up to 4, participated online over a 5-week period in May and June, 2025.

To reduce the annotation effort, participants were presented with \textit{pre-annotated images} using the RT-DETR-X fine-tuned model discussed earlier. Annotators were required to verify and correct these pre-annotations: they could adjust bounding box positions and labels, remove boxes that were incorrect, or add boxes for unannotated vehicles. 
We selected images with the highest disagreement scores for annotation to ensure that the human effort is focused on the most informative and generalizable samples. 
To prevent annotator fatigue and maintain engagement, images of varying difficulty levels were presented to the participant, with the difficulty increasing as they progressed.

\begin{figure}[t!]
    \centering
    \subfloat[Bounding box count per image.]{
        \includegraphics[width=0.45\textwidth]{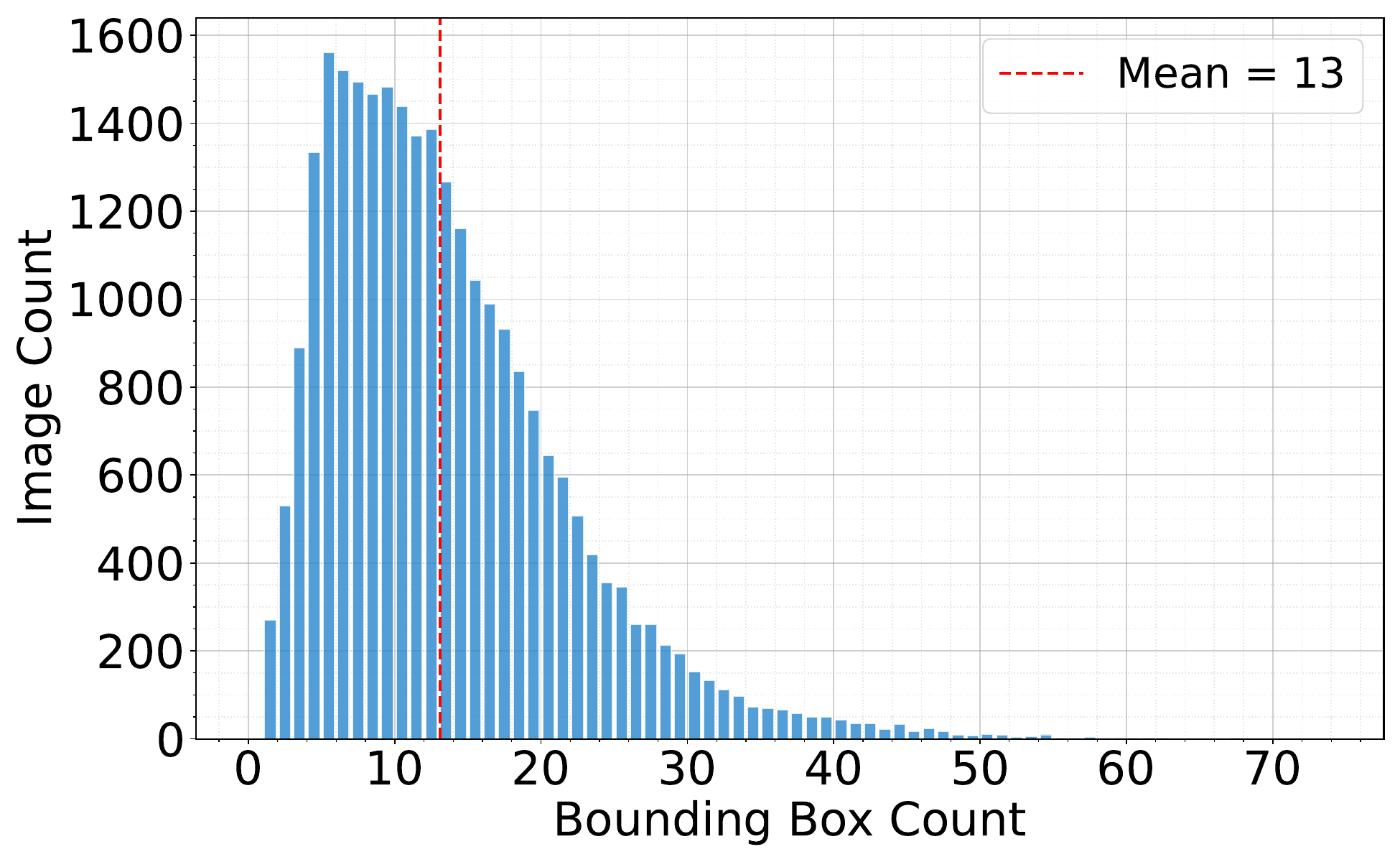}
        \label{fig:bbox_count_per_image_presented_to_users}
    }
    \hfill
    \subfloat[Bounding box area per image ($\times10,000~\text{pixel}^2$).]{
        \includegraphics[width=0.45\textwidth]{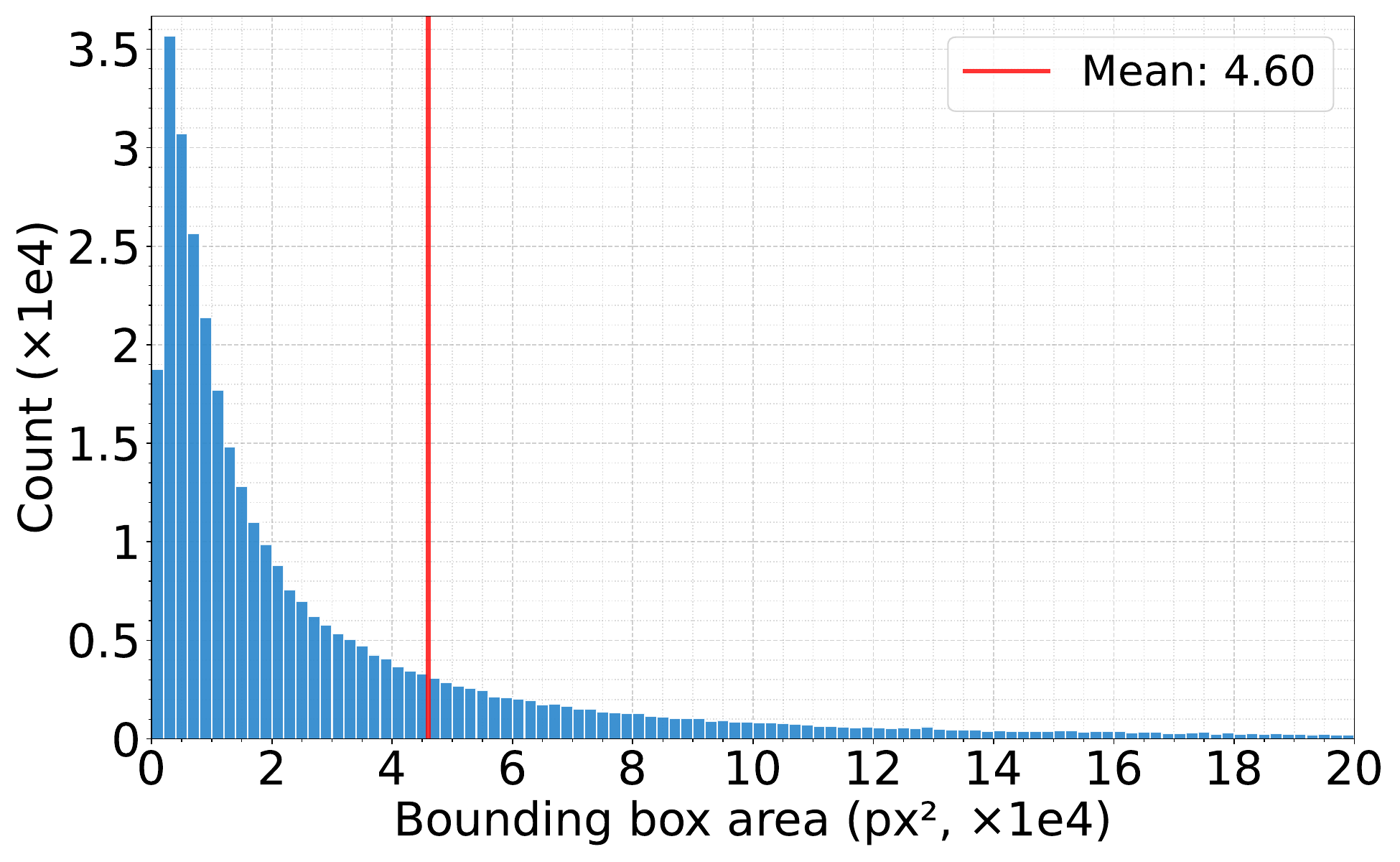}
        \label{fig:area_histogram_presented_to_users}
    }
    \caption{Distribution of bounding box counts and bounding box area per image in pre-annotated images presented to users. This gives a sense of the difficulty of the images being annotated}
    \label{fig:bbox_distribution_presented_to_users}
\end{figure}

Examples of image frames and annotated images are illustrated in Figure~\ref{fig:example_classes_2}.
The overall distribution of pre-annotated bounding boxes presented to participants is shown in Figure~\ref{fig:bbox_distribution_presented_to_users}. The number of bounding boxes per image (Figure~\ref{fig:bbox_count_per_image_presented_to_users}) had a mean count of 13 boxes per image, and the bounding box area by vehicle class (Figure~\ref{fig:area_histogram_presented_to_users}) had a mean area of $460,000~\text{pixel}^2$. These indicate the level of difficulty of the images that the participants were asked to label.

Teams advanced through levels, with their scores and rankings shown in real time. Progression thresholds on both number of images labeled and their accuracy grew stricter with higher levels to ensure that annotation speed does not compensate for low accuracy.
We use the gold images to ascertain the \textit{accuracy} of the participants' labeling. Specifically, 
images were presented to participants in levels with $15$ images each, of which $5$ were gold images with known ground-truth used for accuracy assessment and $10$ were non-gold images for \textit{ab initio} annotation. Participants were not informed that gold images exist, the ordering of gold and non-gold images was randomized, and the gold images were indistinguishable from the non-gold ones.

We solicited \textit{multiple independent annotations per image} for reliability. We assigned each image to participants in a way that maximized both coverage and annotation efficiency. Each participant received a unique set of images, ensuring that no image was repeated within their assigned level. 
This scheduling approach balances annotation quality, participant workload, and dataset coverage, providing a systematic way to gather reliable annotations while maintaining a smooth, gamified annotation workflow. As we discuss in \S~\ref{sec:consensus}, these multiple annotations per image are then used to generate the final consensus annotation for each.

\begin{figure}[t]
        \centering
        \subfloat[Distribution of \# of cumulative bounding boxes per class, across all users and images]{
            \includegraphics[width=0.45\textwidth]{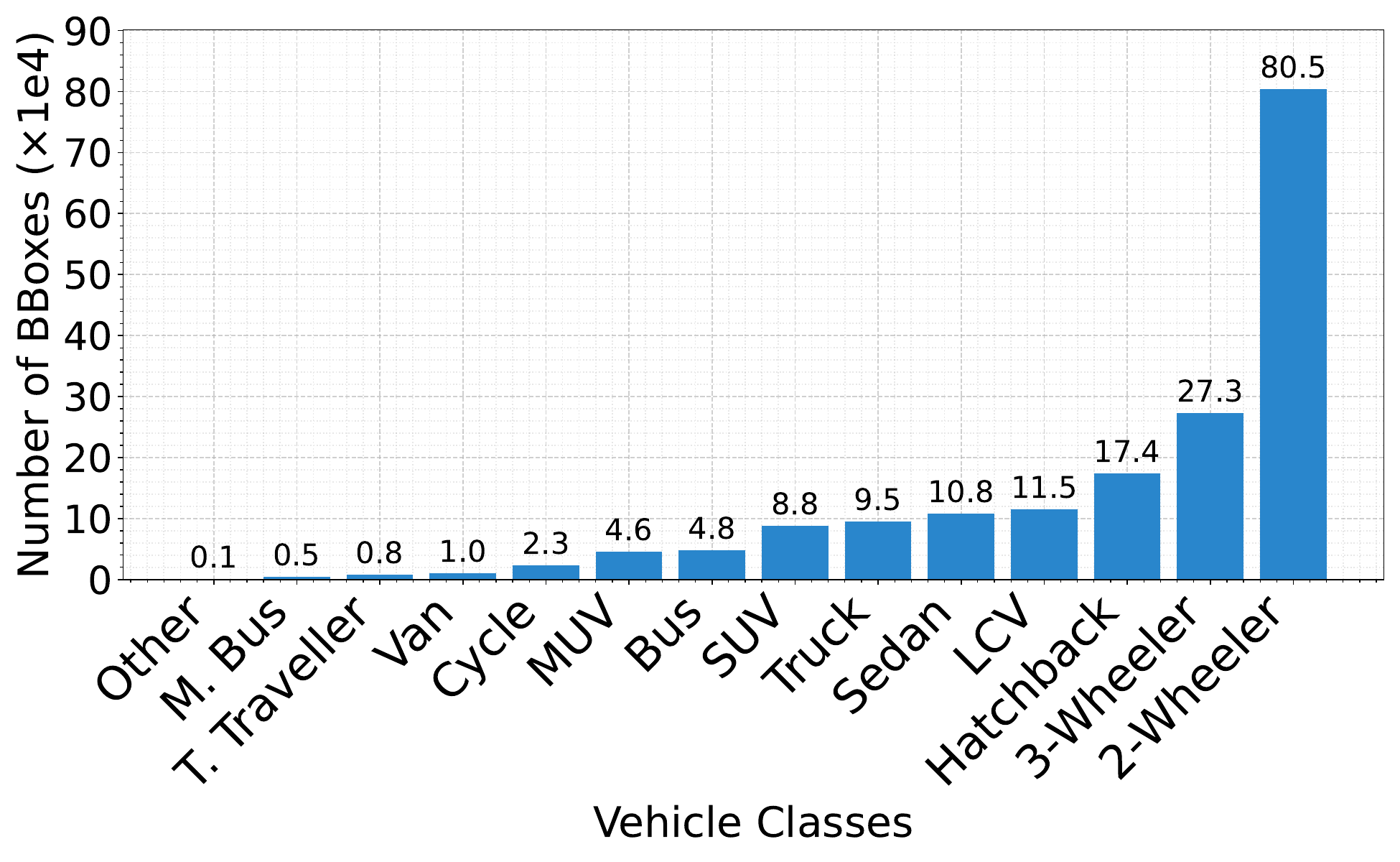}
            \label{fig:vehicle_category_distribution}            
        }
        \hfill
        \subfloat[\# Participants who label the same image]{
            \includegraphics[width=0.45\textwidth]{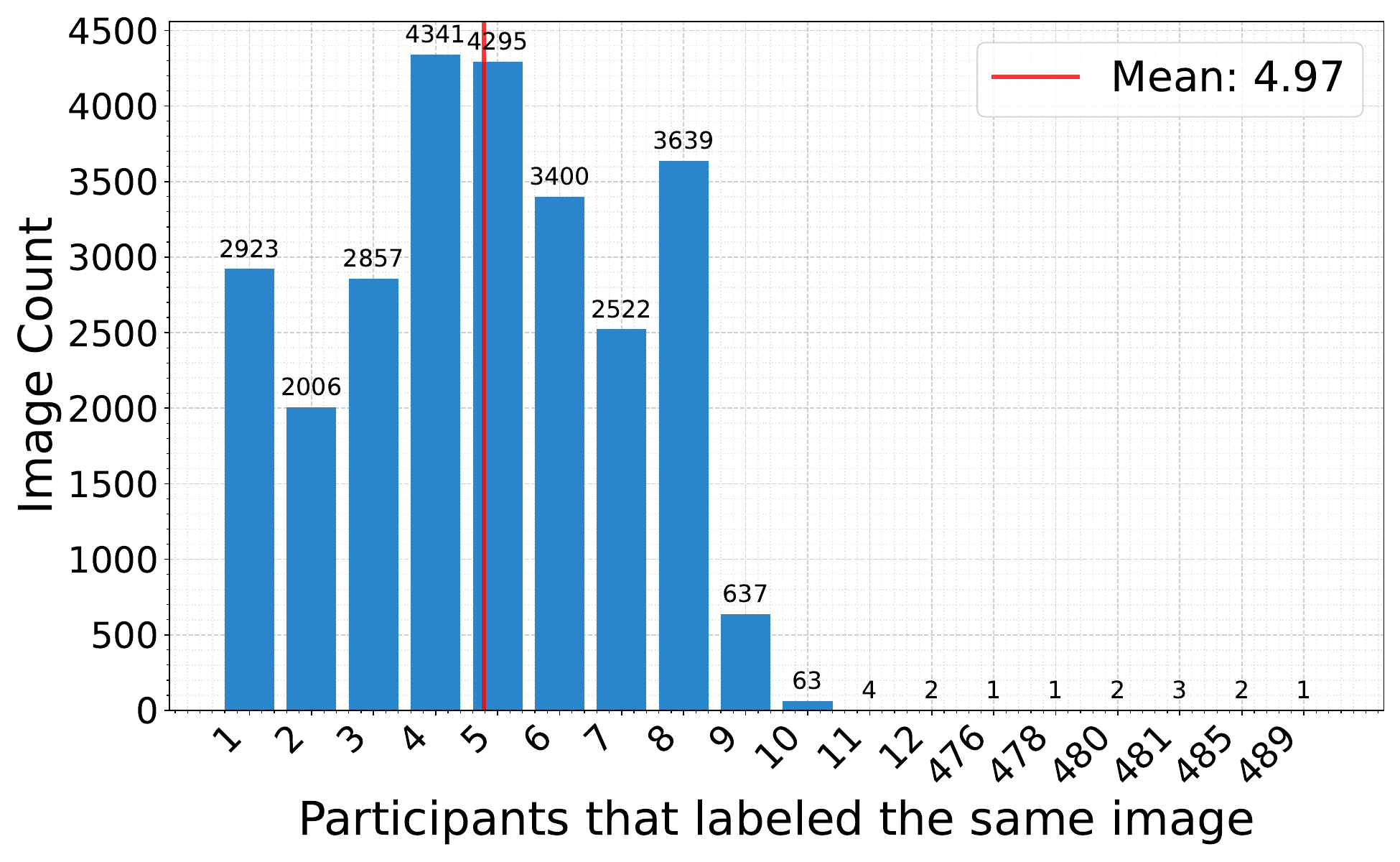}
            \label{fig:user_image_distribution}
        }
        \caption{Distribution of annotations per vehicle classes and number of participants who annotated each image. 
        Each distinct image is annotated by multiple participants to get a total of 1.8M cumulative bounding boxes across the 26k images.}
        \label{fig:data-stats}
\end{figure}

A total of $1,798,324$ bounding boxes were cumulatively annotated across the $26,646$ images. The per-class breakdown is reported in Figure~\ref{fig:data-stats}. Figure~\ref{fig:vehicle_category_distribution} shows the distribution of vehicle classes across the 1.8M bounding boxes, with 2-wheelers, 3-wheelers and hatchbacks being the most common ($>150,000$ boxes), and Figure~\ref{fig:user_image_distribution} illustrates the distribution of participants annotating each image ($4.97$ on average), highlighting the variability in annotation density across the dataset.

\section{Privacy Preservation and Anonymization}
\label{sec: blurring}
We have made a nominal effort to remove personally identifiable information and contextual overlays from the \uvhw images to respect privacy. In particular, we blur vehicle license plates, human faces, and on-frame camera text overlays that include camera identifiers and timestamps. These steps follow established practice in public driving and street-view datasets that apply redaction before release~\cite{frome2009streetview,cityscapes_scripts,waymo_open_faq,mapillary_privacy}. It is important to note that during the annotation phase and hackathon-based data collection, participants were provided with the original, non-blurred CCTV frames to ensure accurate labeling of fine-grained vehicle classes and small objects. The anonymization process described in this section was applied only after the completion of all annotations.

\paragraph{License plates}
Following the anonymization practice adopted in large-scale traffic datasets and open-source anonymizers~\cite{dashcam_anonymizer,waymo_open_faq,mapillary_privacy}, we detect license plates using a YOLO-based one-stage detector trained for road scenes. Detected regions are blurred using a Gaussian kernel with odd dimensions proportional to the plate’s bounding-box size.
To handle varying image resolutions, the detector operates at multiple input scales, ensuring both small and distant plates are masked. This process ensures all alphanumeric content is completely unrecognizable while maintaining a natural image appearance in surrounding regions.

\paragraph{Faces}
Faces are detected using an efficient modern face detector based on the SCRFD architecture~\cite{scrfd2021}. To improve recall under varied illumination and diverse skin tones typical of outdoor CCTV environments, we apply white balance correction, contrast-limited adaptive histogram equalization (CLAHE), gamma adjustment, and unsharp masking prior to detection. Multi-scale and tiled inference is used for high-resolution frames to detect small or partially occluded faces. Each detected face region is expanded by $20\%$ of its bounding-box dimensions and blurred with an adaptive Gaussian kernel.
This ensures that facial details blurred while preserving overall scene context. Prior work suggests that such redaction preserves downstream model performance for person-related tasks while significantly lowering re-identification risk~\cite{dietlmeier2021facesreid}.

\paragraph{On-frame camera overlays}
Text overlays containing camera identifiers, timestamps and location labels are removed through an OCR-driven redaction pipeline. We first apply an OCR model based on PP-OCRv3~\cite{ppocrv3} over predefined regions known to contain overlays --, top-left, top-right, bottom-right, and bottom-left. Detected text polygons are expanded by a fixed pixel margin to ensure full coverage of rendered characters. These masked areas are then removed using the fast-marching inpainting method~\cite{telea2004} implemented in OpenCV~\cite{opencv_bradski2000}, which propagates nearby background pixels to fill the region smoothly. This approach eliminates readable identifiers while preserving spatial consistency and avoids introducing large uniform patches that may bias visual models.

\section{Quality Control using Consensus Algorithms}\label{sec:consensus}

Given that this is a voluntary crowdsourced activity, it is possible that not all participants are performing the annotation tasks with high accuracy. As a result, as discussed above, each image is annotated independently by multiple participants in the UVH challenge.
This requires us to derive a single, high-quality ground truth annotations for each image from these multiple annotations. 
We employ consensus-based aggregation strategies for this\cite{zhou2019crowdsourcing, sheshadri2013square}. Object detection involves two components: the spatial localization of objects (bounding box coordinates) and their categorical assignment (class labels). We treated them separately. 

\subsection{Bounding Box Consensus}
We use a \textit{simple averaging mechanism} for determining the \textit{consensus bounding box} for an object in an image. A practical challenge lies in determining whether bounding boxes from different annotators correspond to the same vehicle/object instance. In our case, this was simplified by the use of pre-annotations: each bounding box shown to annotators carried a persistent identifier, which remained unchanged unless the box was newly added by a participant. Thus, consensus can be directly computed by grouping annotations via their IDs, without repeatedly computing overlaps. 
For newly added bounding boxes, however, no such ID existed since they are not part of pre-annotations shown to annotators; in these cases, we matched boxes across annotators using an IoU threshold of $0.60$ to decide correspondence. Consensus bounding box coordinates were estimated by averaging the submitted annotations across all annotators for a given object instance, thereby smoothing individual biases and capturing a consensus localization. Among all annotations, $\approx 1.64\%$ involved adjustments to bounding box coordinates, $\approx 1.28\%$ represented newly added boxes, $\approx 6.74\%$ corresponded to deletions of pre-existing boxes. When also excluding annotations with label changes (see below), we have $\approx 83\%$ left unchanged.
Since annotators typically made few adjustments to the pre-labeled coordinates, these averaged values are generally very close to those provided by the initial model-assisted annotations. 

\subsection{Class Label Consensus}
For \textit{class labels}, we explored two established consensus techniques: \textit{Majority Voting (MV)}, which assigns the most frequently chosen label to each object, and \textit{Simultaneous Truth and Performance Level Estimation (STAPLE)}~\cite{warfield2004simultaneous}, which jointly models annotator reliability and latent ground truth to produce a probabilistic consensus. Across all annotations, $\approx 7.34\%$ involved a change in the assigned class label compared to the pre-annotations. 
In our experiments, models trained using MV-derived ground truth performed better than those trained with STAPLE-derived annotations, suggesting that the simpler aggregation method yields more consistent supervision under our large-scale, crowd-sourced annotation setting which is consistent with the existing studies~\cite{de2024consensus, heim2018large}. 
We report quantitative comparison between models trained using Majority Voting and STAPLE on \textit{non-anonymized data} in Appendix~\ref{app:anon_datasets:brief}, and will report updated results for models trained on \textit{anonymized data} in a future version of thsi report.

\begin{figure}[t]
    \centering
    \subfloat[Bounding box distribution per class]{
        \includegraphics[width=0.45\textwidth]{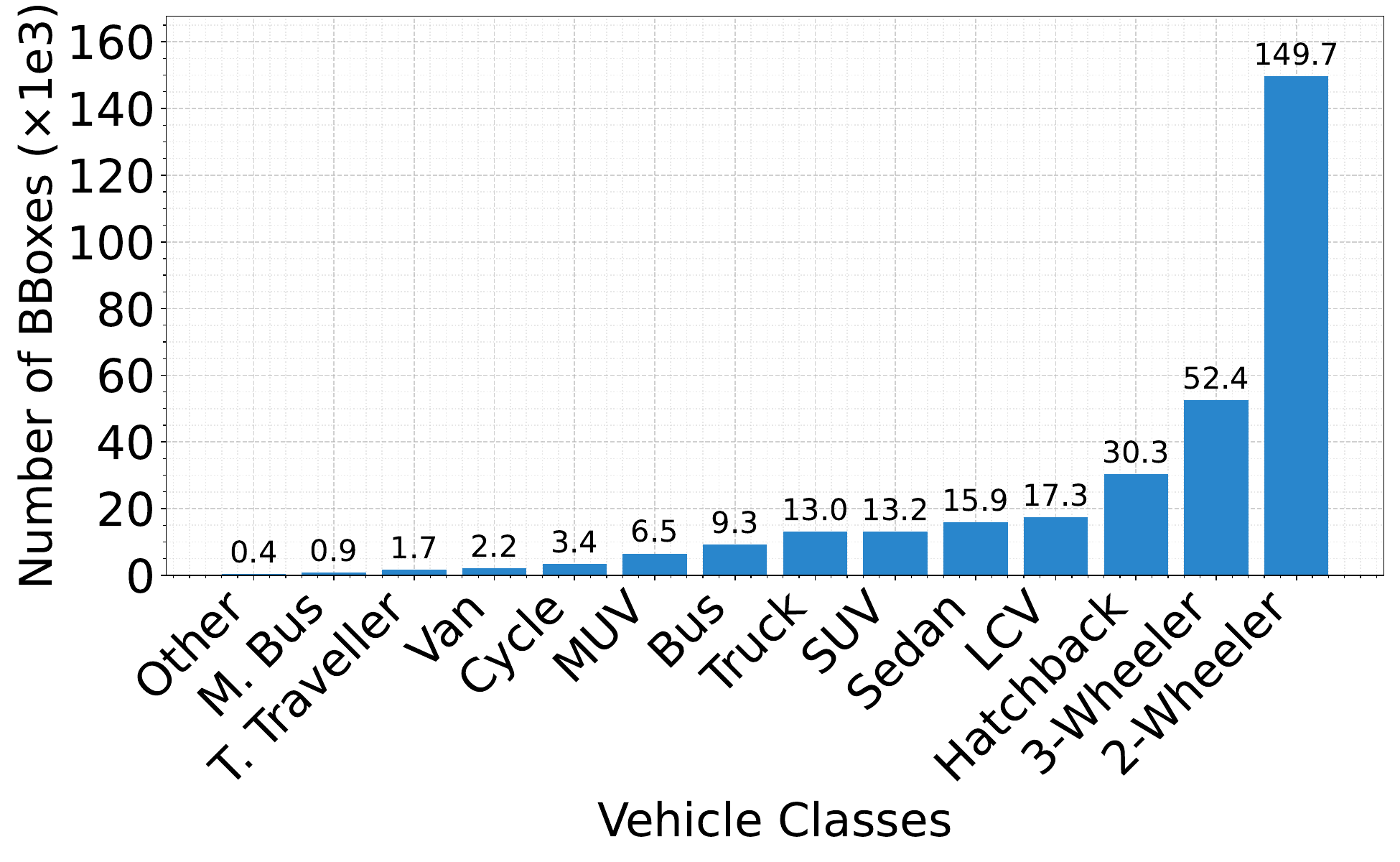}
        \label{fig:instances_per_class_majority_voting}
    }
    \hfill
    \centering
    \subfloat[Bounding box distribution per image]{
        \includegraphics[width=0.45\textwidth]{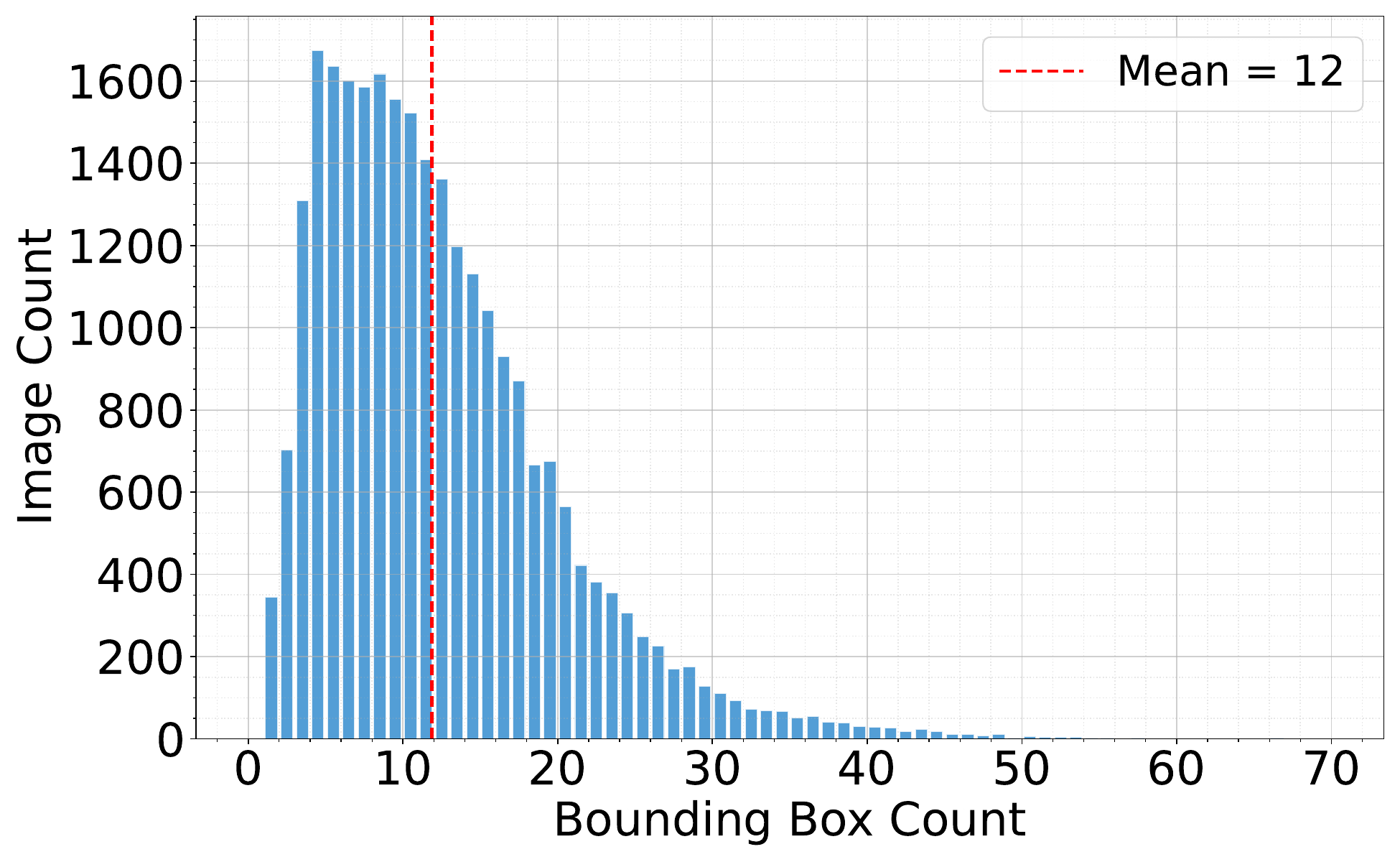}
        \label{fig:bbox_count_per_image_majority_voting}
    }
    \caption{Distribution of Bounding box counts in \uvhmv consensus dataset.}
    \label{fig:bbox_count_majority_voting}
\end{figure}

\begin{figure}[t]
    \centering
    \includegraphics[width=0.45\textwidth]{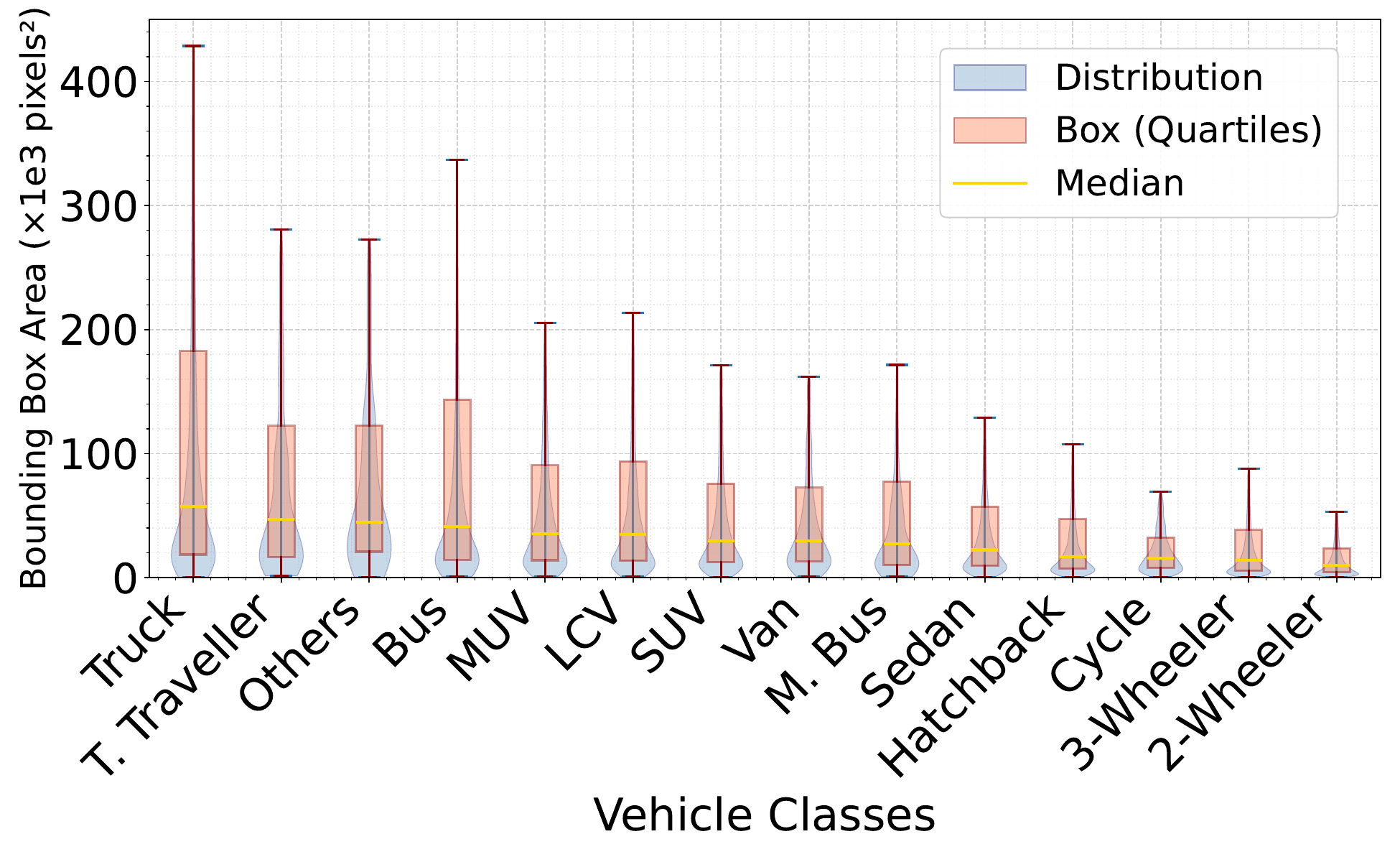}
    \caption{Distribution of Bounding box area per class in in \uvhmv consensus datasets.}
    \label{fig:area_distribution_per_class_majority_voting}
\end{figure}

\subsubsection{Majority Voting} 

In the Majority Voting (MV) approach, each distinct bounding box in a pre-labeled image is assigned a class label based on the most frequently selected label among all annotators who contributed to that box. In the rare case of a tie, where two or more vehicle classes receive an equal number of votes, the final label is chosen randomly from among them. 
Of the 1.8M individual bounding boxes contributed by annotators, applying Majority Voting resulted in $316,220$ distinct consensus bounding boxes across the $26,646$ images. This dataset with the \uvh images and annotations using MV are referred to as \texttt{\uvhmv}.
The resulting consensus bounding box statistics are summarized in Figure~\ref{fig:instances_per_class_majority_and_staple}, which includes the class-wise distribution of consensus bounding boxes (Figure~\ref{fig:instances_per_class_majority_voting}) and the distributions of bounding box counts per image (Figure~\ref{fig:bbox_count_per_image_majority_voting}), and bounding box area per class can be seen in Figure~\ref{fig:area_distribution_per_class_majority_voting}.

\subsubsection{STAPLE}\label{sec:app-staple}

\begin{algorithm}[t]
\caption{Adapted STAPLE for Object Detection}
\label{alg:staple}
\KwIn{
  $M \in \mathbb{Z}^{N_a \times N_b}$: annotation matrix (annotators $\times$ bboxes); \\
  $S \in [0,1]^{N_a \times C}$: per-class sensitivity; \\
  $T \in [0,1]^{N_a \times C}$: per-class specificity; \\
  $C$: number of classes; $I$: max iterations; $\epsilon$: convergence threshold.
}
\KwOut{
  $\hat{y} \in \{1,\dots,C\}^{N_b}$: consensus labels
}

\BlankLine
Initialize class prior $\pi_k = 1/C$ for all $k \in \{1,\dots,C\}$\;
Initialize annotator reliability matrix $\theta_{j}$ from $S,T$\;

\For{$t=1$ \KwTo $I$}{
  \tcp{E-step: posterior distribution over true labels}
  \For{each bbox $i=1 \dots N_b$}{
    \For{each class $k=1 \dots C$}{
      Compute $\log W_{i,k} = \log \pi_k + \sum_j \log \theta_{j}(M_{j,i}, k)$\;
    }
    Normalize $W_{i,:}$ to sum to 1\;
  }

  \tcp{M-step: update parameters}
  Update priors $\pi_k = \frac{1}{N_b}\sum_i W_{i,k}$\;
  Update annotator reliabilities $\theta_j$ using $M$ and $W$\;

  \If{$\|\theta^{(t)} - \theta^{(t-1)}\|_\infty < \epsilon$}{break\;}
}
$\hat{y}_i = \arg\max_k W_{i,k}$
\Return $\hat{y}$
\end{algorithm}

Simultaneous Truth and Performance Level Estimation (STAPLE)~\cite{warfield2004simultaneous} is an iterative Expectation–Maximization (EM) algorithm originally introduced for medical image segmentation, where the goal is to estimate a latent ``true'' segmentation from multiple noisy annotators. In our setting, we adapt STAPLE to the object detection domain with the idea of jointly estimating both the consensus ground truth and the reliability of individual annotators.

At each iteration, the algorithm computes per-class sensitivity (true positive rate) and specificity (true negative rate) for every annotator, thereby modeling their likelihood of correctly labeling an object of a given class. 
This is done based on the running accuracy of the annotators in labeling the gold images in each level.
These parameters are then used to re-weight the annotators' contributions when inferring the consensus bounding box labels. In contrast to simple majority voting, which assumes all annotators are equally reliable, STAPLE explicitly down-weights the influence of lower quality annotators while giving more weight to consistent and accurate annotators. This makes STAPLE potentially effective in reducing the impact of noisy labels and systematic annotator biases, yielding a more robust consensus ground truth. 

\begin{figure}[t!]
    \centering
    \subfloat[Bounding box distribution per class]{
        \includegraphics[width=0.45\textwidth]{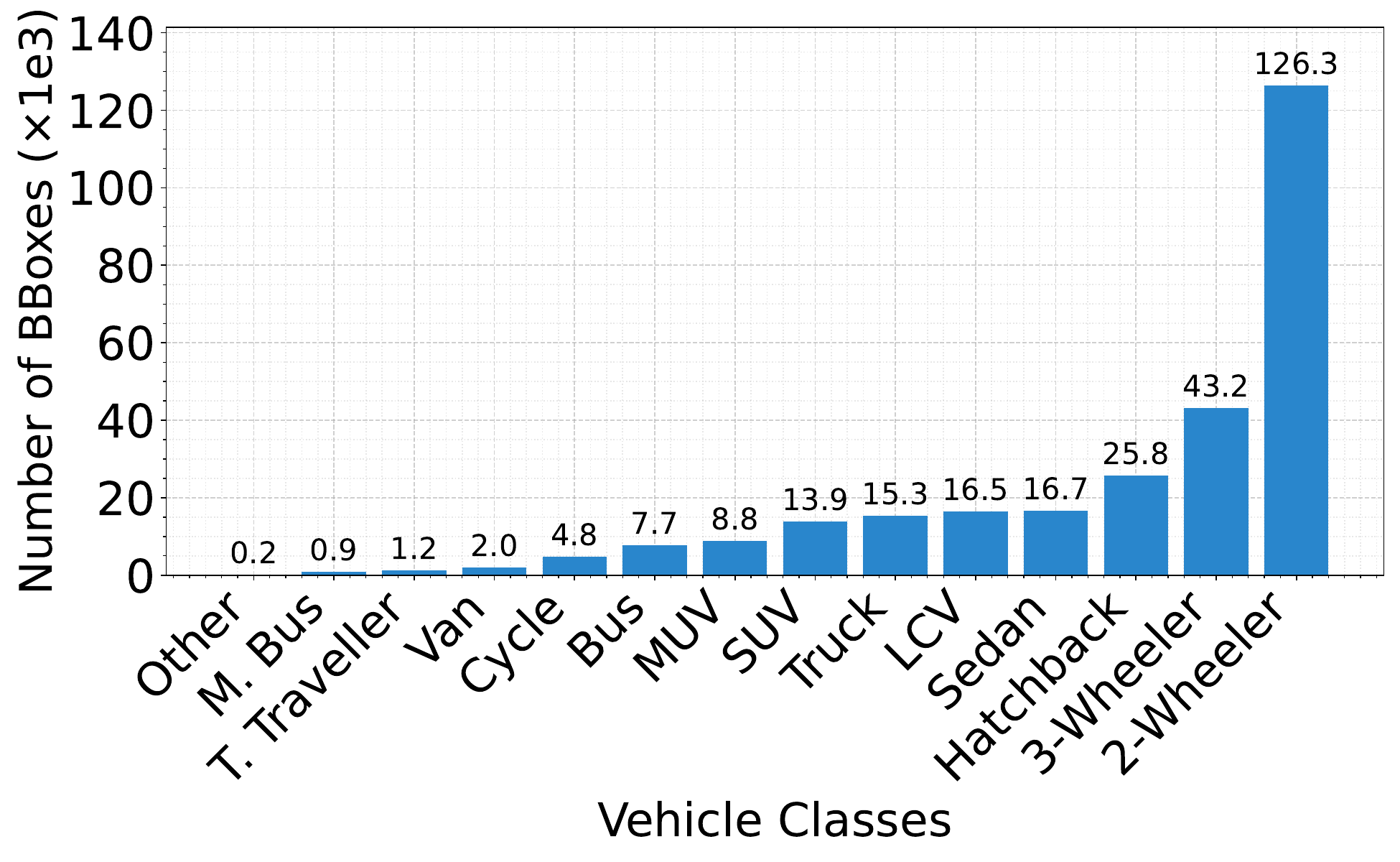}
        \label{fig:app-instances_per_class_staple}
    }
    \hfill
    \centering
    \subfloat[Bounding box distribution per image]{
        \includegraphics[width=0.45\textwidth]{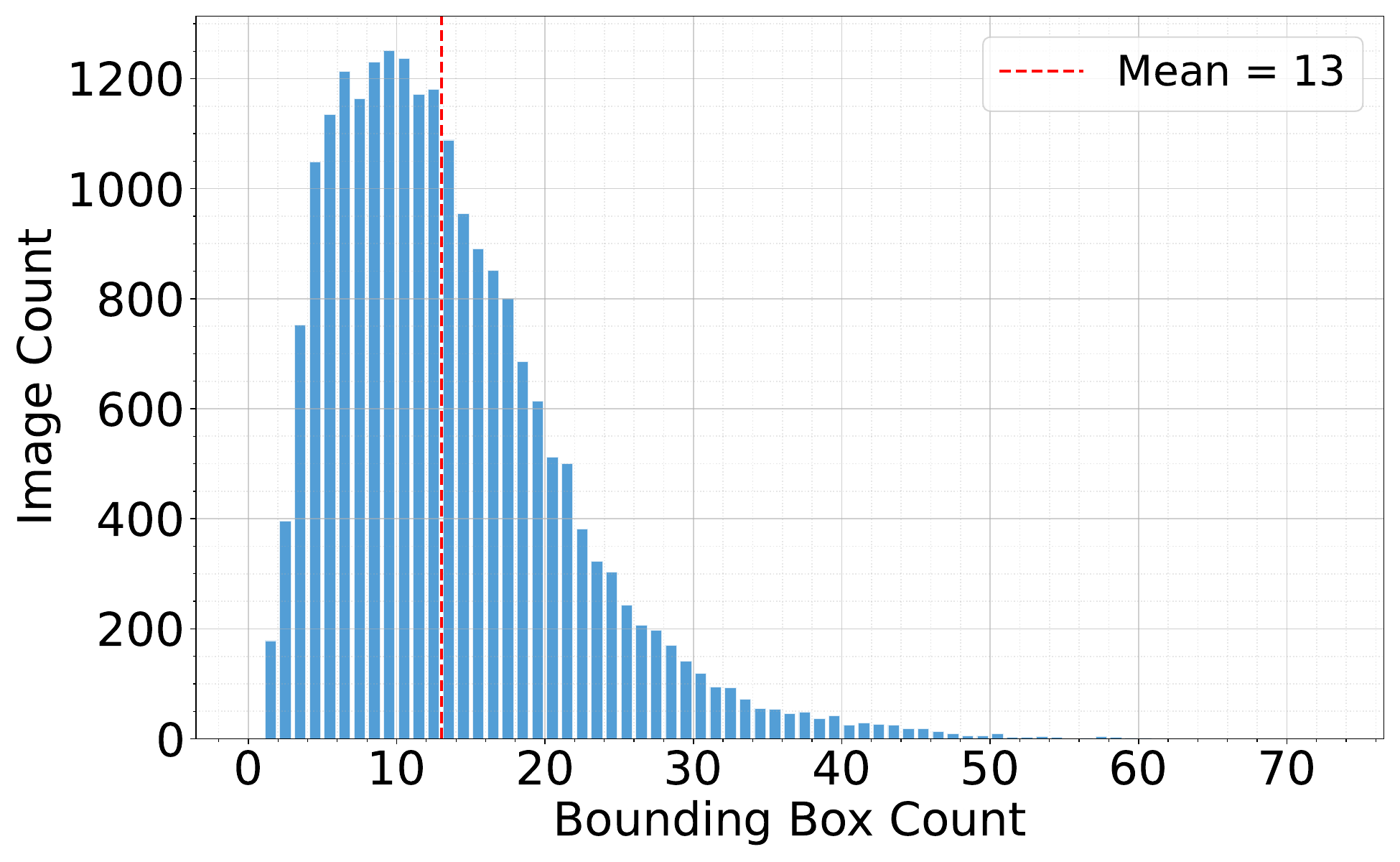}
        \label{fig:app-bbox_count_per_image_staple}
    }
    \caption{Distribution of Bounding box counts in \uvhst consensus dataset.}
    \label{fig:instances_per_class_majority_and_staple}
\end{figure}

\begin{figure}[t]
    \centering
    \includegraphics[width=0.45\textwidth]{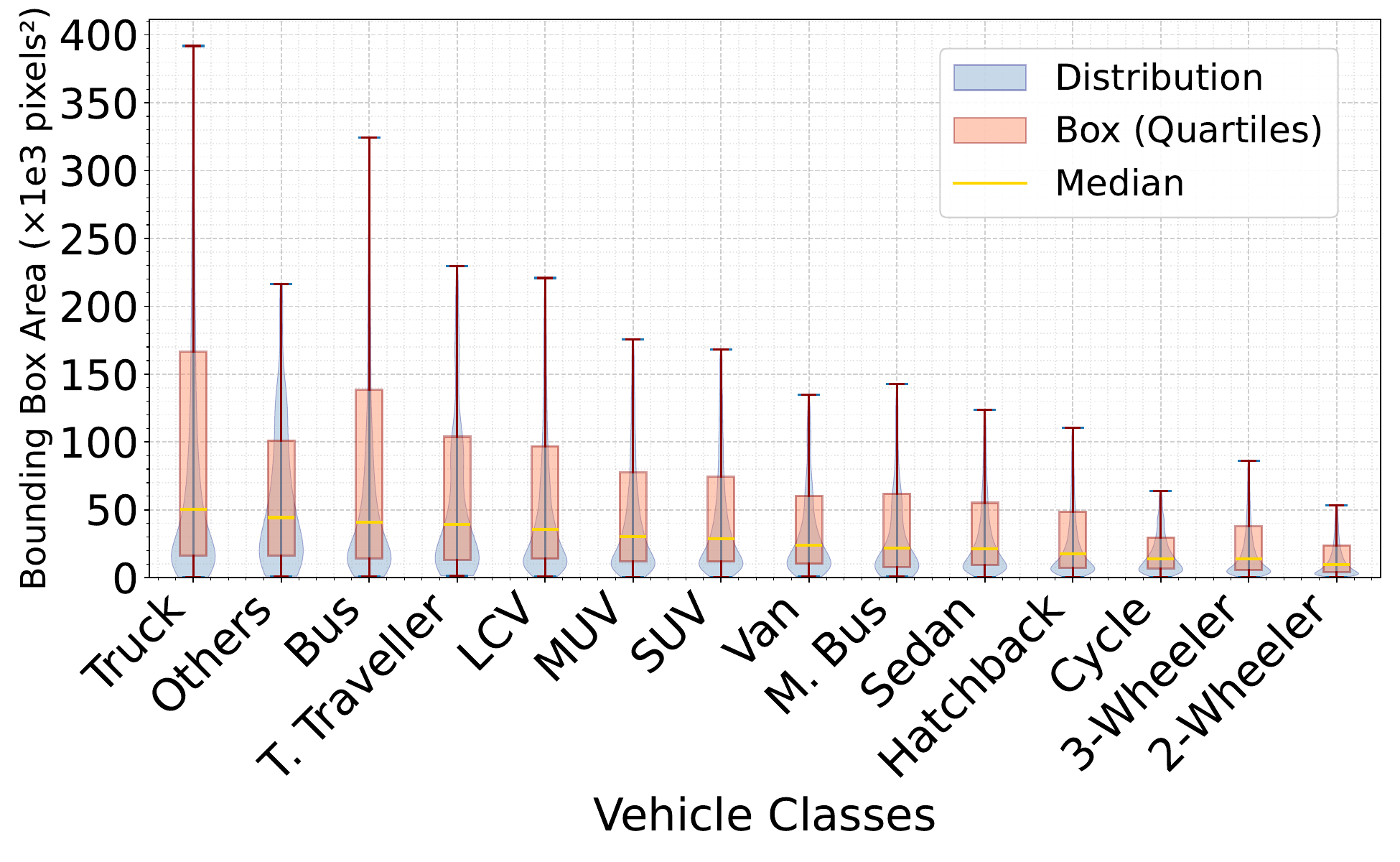}
    \caption{Distribution of Bounding box area per class in in \uvhst consensus datasets.}
    \label{fig:app-area_distribution_per_class_staple}
\end{figure}

Algorithm~\ref{alg:staple} gives the high-level pseudocode for the STAPLE algorithm, adapted by us for object detection consensus based on the participants' annotations.
The algorithm takes as input an annotation matrix $M$ representing the class labels assigned by $N_a$ annotators to $N_b$ bounding boxes for a given image, along with each annotator’s estimated per-class sensitivity $S$ and specificity $T$. The number of classes is denoted by $C$, and the iterative EM process runs for at most $I$ iterations or until the convergence threshold $\epsilon$ is reached. In each iteration, the E-step estimates the posterior probability of the true class for every bounding box based on annotator reliability, while the M-step updates the class priors and annotator reliability parameters. The procedure converges when the estimated reliabilities stabilize, yielding the consensus class label $\hat{y}$ for each bounding box.

We refer to this \uvh dataset with STAPLE annotations as \texttt{\uvhst}.
Using STAPLE, the 1.8M raw bounding box annotations were aggregated into $283,402$ consensus bounding boxes across the 26k images. The per-class distribution of these consensus bounding boxes is presented in Figure~\ref{fig:app-instances_per_class_staple}, the distribution of bounding box counts per image is in Figure~\ref{fig:app-bbox_count_per_image_staple}, and the distribution of bounding box area per class can be seen in Figure~\ref{fig:app-area_distribution_per_class_staple}.

\section{Model Fine-tuning using \uvhw Dataset}\label{sec:train}

Models presented in this section are fine-tuned by training using the \textbf{\uvhmv consensus annotations} (\S~\ref{sec:consensus}) since these offered better accuracy than those trained using \uvhst. 
The models placed in the public domain at this time are also trained on \uvhmv.
In the near future, we will report results for models trained on \uvhst and place those models in the public domain as well. All training and evaluation are performed on the \textit{anonymized dataset} described in Section~\ref{sec: blurring}. However, for completeness, we also include in Appendix~\ref{app:anon_datasets:brief} the results of models trained on the non-anonymized images, though those images and models trained on them will not be made public for privacy reasons.

\begin{table}[t]
    \centering
    \caption{Model specifications and compute requirements for the retraining object detection models.}\label{tab:model_specifications}
    \small
    \begin{tabular}{lcrr}
        \toprule
        \bf Model & \bf Input Size & \bf Parameters (M) & \bf FLOPs (G) \\
        \hline\hline
        DAMO-YOLO-T\cite{xu2022damo} & 640×640 & 8.5 & 18.1 \\
        DAMO-YOLO-L\cite{xu2022damo} & 640×640 & 42.1 & 97.3 \\
        YOLOv8-S\cite{jocher2023yolov8} & 640×640 & 11.2 & 28.6 \\
        YOLOv8-X\cite{jocher2023yolov8} & 640×640 & 68.2 & 257.8 \\
        YOLOv11-S\cite{jocher2024yolov11} & 640×640 & 9.4 & 21.5 \\
        YOLOv11-X\cite{jocher2024yolov11} & 640×640 & 56.9 & 194.9 \\
        RT-DETRv2-S\cite{zhao2024rtdetr} & 640×640 & 20 & 60 \\
        RT-DETRv2-X\cite{zhao2024rtdetr} & 640×640 & 76 & 259 \\
        \bottomrule
    \end{tabular}    
    \label{tab:model-specs}
\end{table}

We use a diverse set of SOTA object detection models that have already been trained on COCO dataset and subsequently fine-tune them on \uvhmv.
These model families represent a range of modern detectors, including the YOLO series (YOLOv11-X/S, DAMO-YOLO-T/L) and transformer-based detectors (RT-DETRv2-X/S). These models were chosen not only for their strong benchmark performance and widespread adoption within the literature but also to represent a balance of accuracy, computational efficiency, and inference speed.

Specifically, we fine-tune models from the YOLO family of fast and lightweight detectors that are amenable to real-time application even on edge devices. 
We select DAMO-YOLO from Alibaba available under the flexible Apache 2.0 license, and YOLOv11-S from Ultralytics provided under the more restrictive AGPL-3.0 License. We also fine-tune RT-DETRv2 from Baidu, available under the Apache 2.0 license, which highlights the latest advancements in transformer-based architectures designed to enhance accuracy under challenging detection conditions. 
Table~\ref{tab:model_specifications} summarizes the core model specifications, including input resolution, parameter count, and FLOPs.

We construct a single stratified training-validation split of the \uvhw consensus dataset, using $80\%$ of the $26,646$ images for training (\uvhw-Train) and $20\%$ for validation (\uvhw-Val). Randomization was introduced during the split generation to reduce sampling bias.
All training was conducted using the PyTorch Python framework on NVIDIA A6000, A100 and H200 GPUs from our KIAC GPU Cluster, providing sufficient memory and throughput for large-scale experiments. We start with initial model weights for these model architectures trained on the COCO dataset and fine-tune them till convergence.
Batch sizes were set to 16 for all the models. The training pipeline adhered to default hyperparameters of the respective models: the AdamW optimizer for all models, the cosine learning rate scheduler for the YOLO family of models, and the MultiStepLR scheduler for the RT-DETR models. The specific hyperparameters used are provided in Table~\ref{tab:training_hparams} of Appendix~\ref{app:hyperparam}.

\section{Benchmark Experiments}\label{sec:benchmarks}

\subsection{Evaluation Protocol and Metrics}
\label{sec:evaluation-protocol}

\begin{figure}[t]
        \centering
        \subfloat[Instance count per class.]{
            \includegraphics[width=0.45\textwidth]{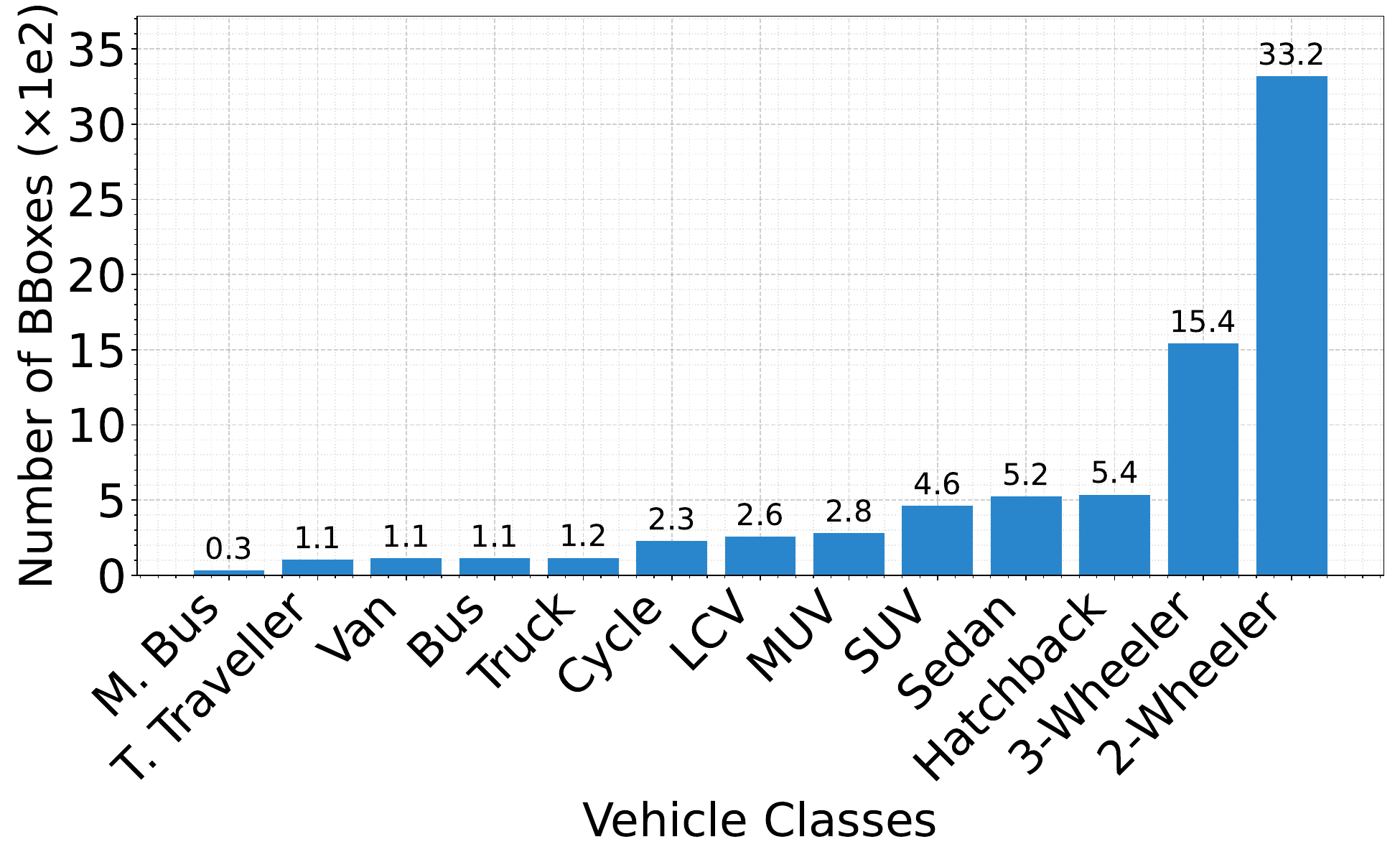}
            \label{fig:instances_per_class}
        }
        \hfill
        \subfloat[Bounding box count per image.]{
            \includegraphics[width=0.45\textwidth]{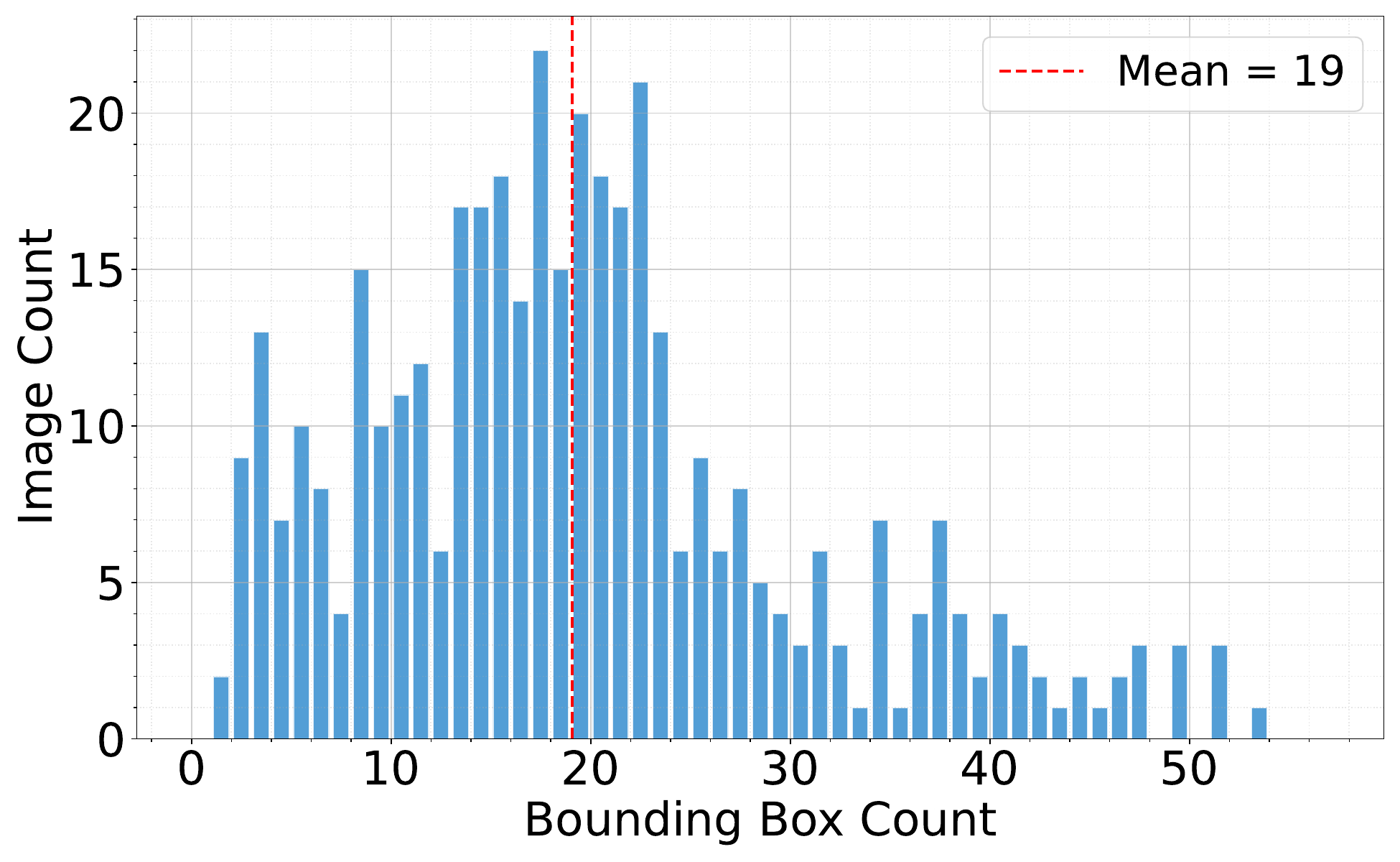}
            \label{fig:bbox_count_per_image}
        }
        \caption{Distribution of \textit{held out test set} from gold dataset used for reporting accuracy metrics. These are not used during fine-tuning.}
        \label{fig:distribution_classes_bbox}
\end{figure}

We evaluate the performance of our fine-tuned model on the \textbf{\uvhmv dataset (anonymized)} and compare it against the corresponding baseline models. 
Results for models trained on \uvhst will be reported in a future version.

We use a \textit{held-out test set} curated from our gold dataset comprising of 400 images, sampled to ensure diverse coverage of all fourteen \uvhw vehicle classes. Figure~\ref{fig:distribution_classes_bbox} summarizes the evaluation dataset, illustrating the per-class instance counts in Figure~\ref{fig:instances_per_class} and the distribution of bounding-box counts per image in Figure~\ref{fig:bbox_count_per_image}. None of these images were used during model fine-tuning.

When comparing our models against the baseline ones trained on the COCO dataset, we identified the subset of vehicle categories that overlap with our class taxonomy. Specifically, we mapped our \uvh classes into three broad categories in COCO: (1) COCO:\emph{Car}, consolidates \uvh:\emph{Hatchback}, \emph{Sedan}, \emph{SUV}, \emph{MUV}, and \emph{Van}; (2) COCO:\emph{Bus}, includes \uvh:\emph{Bus} and \emph{Mini Bus}; and (3) Coco:\emph{Truck} directly maps to \uvh:\textit{Truck}. Although \emph{Cycle} and \emph{2-Wheeler} are present in both datasets, we excluded them from this comparative analysis because of a fundamental annotation mismatch: in COCO, the bounding boxes of these vehicle classes are annotated without the riders, whereas in our dataset, the bounding boxes encapsulate both the vehicle and rider. We have also excluded the ``Others'' class from the evaluation since it is just an umbrella class for all the vehicles that do not belong to any of the 13 primary classes and also have very few instances.

Performance assessment follows standard practices widely adopted in the object detection literature~\cite{lin2014coco, everingham2010pascal}. The primary metric is the \textit{mean Average Precision (mAP)}, evaluated across a range of Intersection over Union (IoU) thresholds. In particular, we report:
\begin{enumerate}
  \item \textit{mAP(50:95)}: The main benchmark metric, defined as the mean of AP values at IoU thresholds from \(0.50\) to \(0.95\) in steps of \(0.05\).  
  \item \textit{mAP(75):} AP computed at a stricter IoU threshold of \(0.75\), which emphasizes precise localization quality.
  \item \textit{mAP(50):} AP computed at a lenient IoU threshold of \(0.50\), reflecting the model's capacity for coarse but correct detections.
\end{enumerate}
More details of these metrics are defined in Appendix~\ref{app:map}.

In addition to the overall performance measures, we also provide \textit{per-class Average Precision (AP)} values to emphasize differences in performance among various vehicle categories. This evaluation approach ensures a comprehensive assessment of the model's capabilities and allows for direct comparisons with existing large-scale benchmarks.

\subsection{Model Performance}
Detailed experimental results are provided in Appendix~\ref{app:anon_datasets}, in Tables~\ref{tab:uvh_per_class_metrics_blurred} to~\ref{tab:hackathon_overall_map_blurred}. We summarize two key aspects below. First, we demonstrate the benefit of fine-tuning models on our proposed \uvhw dataset compared to the baselines using COCO-pretrained weights, for overlapping vehicle classes. Second, we illustrate the relative performance differences across model architectures using various metrics.

\begin{figure}[t!]
    \centering
    \includegraphics[width=0.7\textwidth]{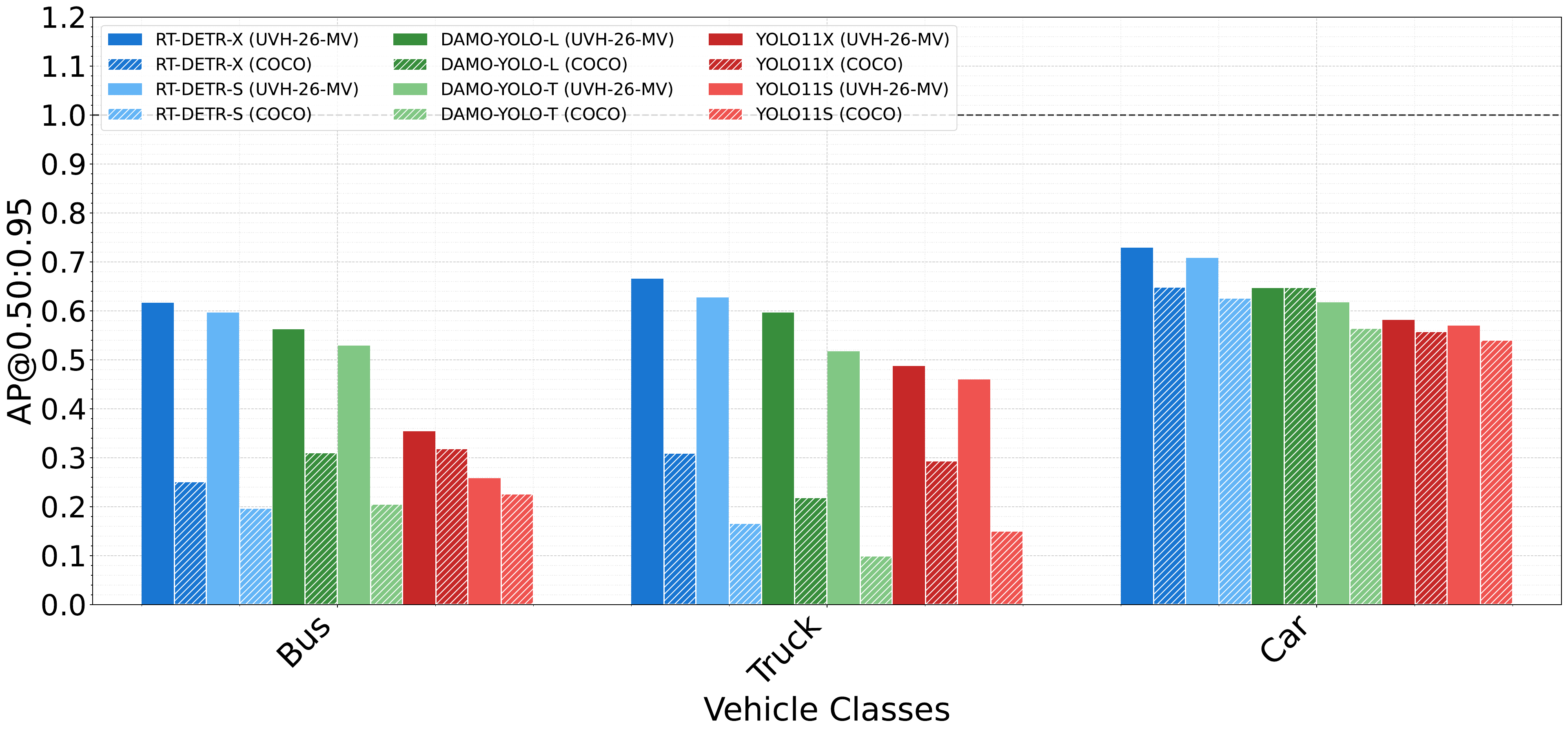}
    \caption{AP@50:95 performance comparison across different models for the common vehicle classes (Car, Bus, Truck). Models fine-tuned on \uvhmv are in solid color while baseline ones trained only on COCO are hatched.}
    \label{fig:ap_50_95_performance_2_blurred}
\end{figure}

\subsubsection{Comparing Models Trained on \uvhw with Baselines Trained on COCO}
Figures~\ref{fig:ap_50_95_performance_2_blurred} and \ref{fig:hackathon_ap_50_95_performance_blurred} show major improvements when models pre-trained on COCO are fine-tuned on \uvhmv, confirming the presence of a strong domain gap between ego-view datasets and top-down CCTV imagery. 

Models trained directly on COCO show limited ability to generalize to the surveillance viewpoint, where vehicles appear smaller, more occluded, and captured from higher elevations. Fine-tuned models on \uvhmv consistently achieve higher $mAP(50:95)$ across all overlapping classes (bus, car, truck). The largest gains are observed for vehicles such as buses and trucks, whose appearance differs most between ego-view and aerial perspectives. These improvements demonstrate that pre-training on general-purpose datasets like COCO provides transferable low-level features, but fine-tuning on contextually aligned data such as \uvhw is critical for high-level scene understanding under surveillance viewpoints.

\subsubsection{Performance of Different Models when Trained on \uvhmv Dataset}
\begin{figure}[t]
    \centering
    \subfloat[Comparison of baseline COCO models and \uvhmv models for classes common to both.]{\includegraphics[width=0.45\textwidth]{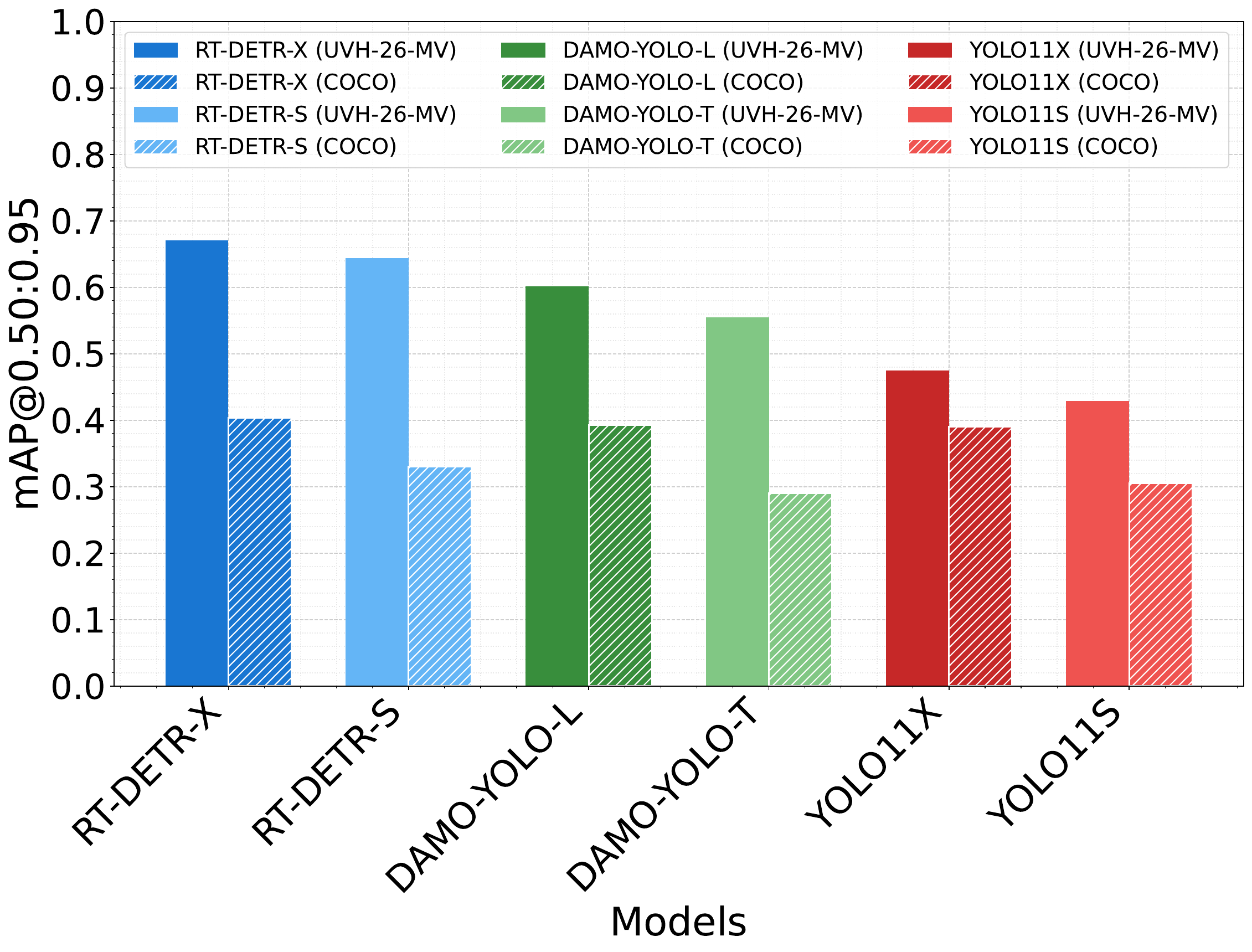}\label{fig:hackathon_ap_50_95_performance_blurred}}
        \hfill
    \subfloat[Results of \uvhmv models on all \uvh classes.]{
    \includegraphics[width=0.45\textwidth]{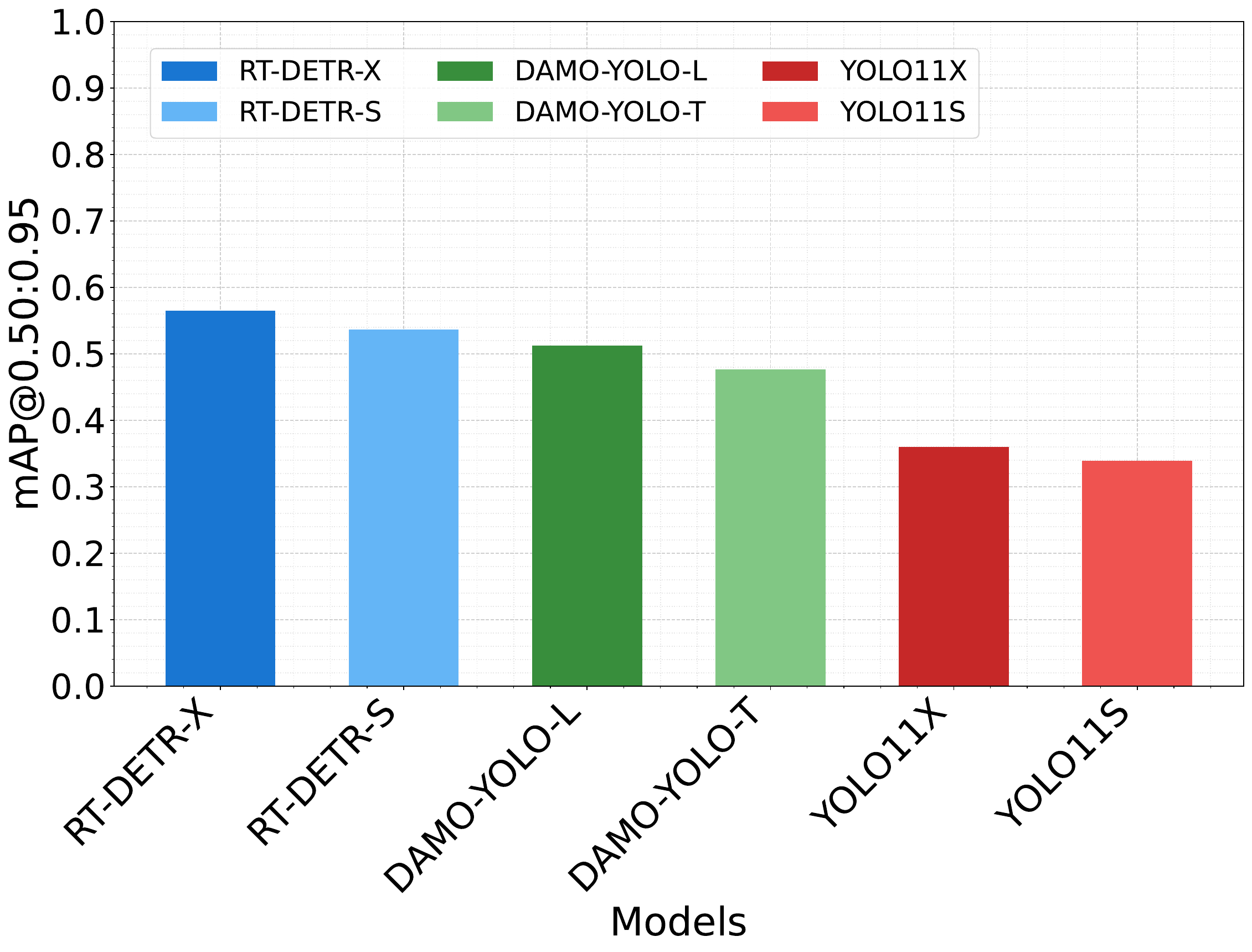}
    \label{fig:hackathon_ap_50_95_performance_2_blurred}}
    \caption{Performance comparison (mAP@50:95)
    of all model architectures.}
    
    \label{fig:difficulty}
\end{figure}

Across all evaluated detectors, transformer-based models outperform convolutional architectures on the \uvhmv dataset, reflecting their stronger capacity to capture long-range dependencies and dense spatial layouts. Both RT-DETR variants achieve the highest mean average precision, while larger YOLO and DAMO-YOLO models provide comparable accuracy. Figure~\ref{fig:difficulty} provides an overall comparison of the performance of these model architectures, contrasting their results on both the common and full set of \uvhmv classes.

\begin{figure}[t]
    \centering
    \includegraphics[width=0.9\textwidth]{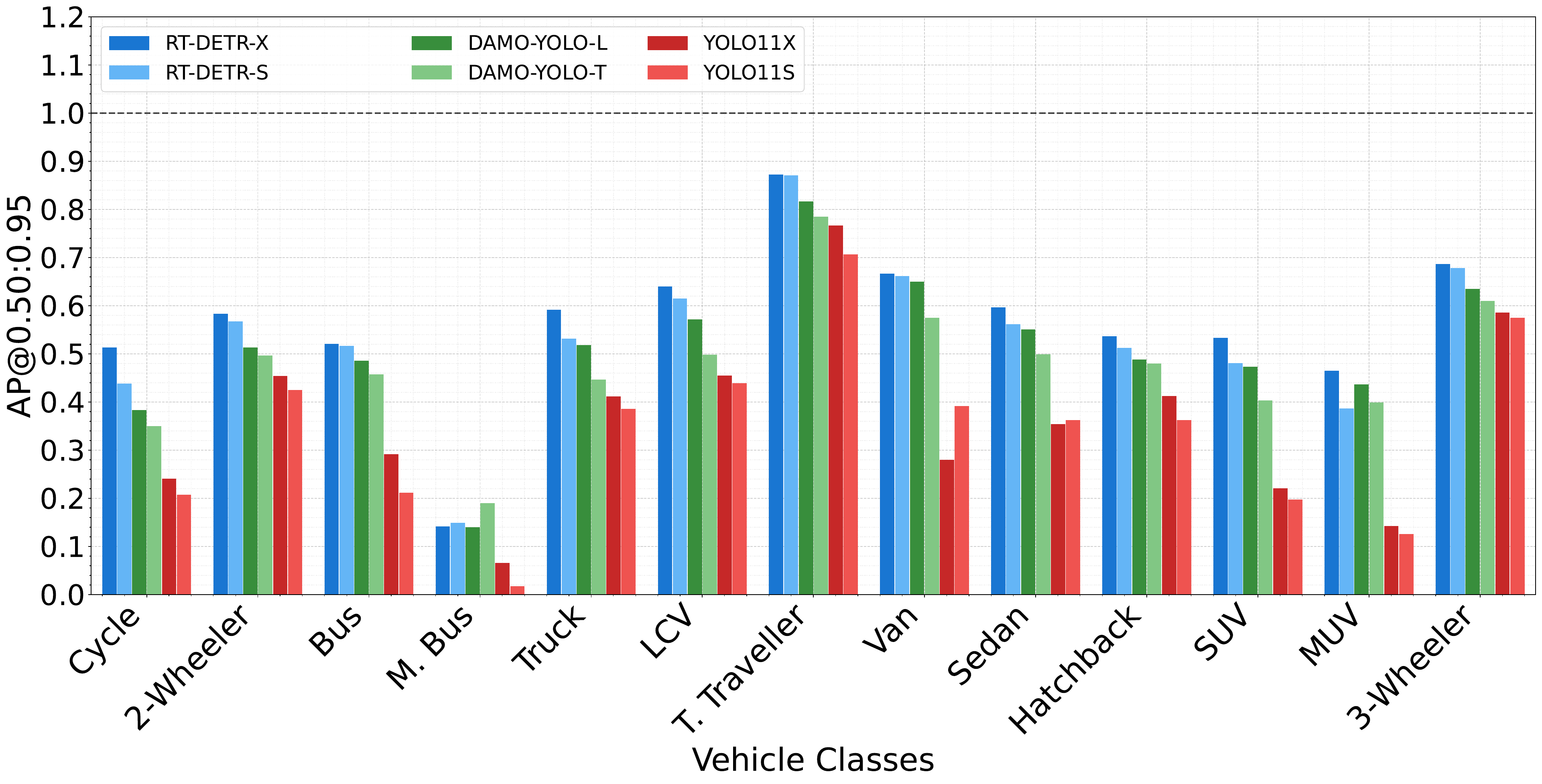}
    \caption{Performance comparison (AP@50:95) of models fine-tuned on \uvhmv, across different model architectures for each \uvh vehicle class.}
    \label{fig:ap_50_95_performance_3_blurred}
\end{figure}

Classes such as Tempo-traveller and Van, although relatively underrepresented, were detected accurately due to their distinctive visual features. In contrast, classes like Mini-Bus exhibited lower detection performance, likely because of limited representation (see Figure \ref{fig:ap_50_95_performance_3_blurred}) and strong visual similarity to Bus. Well-represented and visually distinctive classes, including Two-wheeler and Three-wheeler, were detected reliably. However, fine-grained car classes such as Hatchback and Sedan showed lower detection accuracy despite high representation, primarily owing to their visual similarity to other car subtypes. Figures~\ref{fig:hackathon_ap_50_95_performance_2_blurred} and \ref{fig:ap_50_95_performance_3_blurred} illustrate that the \uvhmv dataset enables a strong adaptation and evaluation of detectors under realistic urban surveillance conditions. Additional details are available in Tables \ref{tab:hackathon_per_class_merged_blurred} and \ref{tab:hackathon_overall_map_blurred} of Appendix~\ref{app:anon_datasets}.

\subsection{Additional Results}
While we are not placing the non-anonymized \uvhw dataset in the public domain, we fine-tune the baseline models using the non-anonymized \uvhw dataset to compare the difference in accuracy between models fine-tuned with and without anonymization. Appendix~\ref{app:anon_datasets:brief} reports results these. It also compares models fined-tuned using \textit{Majority Voting} against \textit{STAPLE} consensus over the non-anonymized datasets. While these results are reported for academic benefit, for privacy reasons, \textbf{we are not releasing} the non-anonymized datasets or models in the public domain.

\section{Data Release and Licensing}
This is the Version 1 release of the anonymized \textit{\uvhw dataset and annotations} and {models} fine-tuned on them, corresponding to the Week 1 of the hackathon. This includes two consensus variants -- the \texttt{\uvhmv} dataset derived using the Majority Voting algorithm and the \texttt{\uvhst} dataset derived using the STAPLE algorithm. We also provide models fine-tuned on the anonymized \texttt{\uvhmv} dataset; models trained on \texttt{\uvhst} dataset will be released in the near future.
The release package also includes utility scripts for data conversion, visualization, and baseline model evaluation.
These datasets are distributed under the \textit{Creative Commons Attribution 4.0 International License}~\footnote{https://creativecommons.org/licenses/by/4.0/}, while the accompanying models, scripts, and source code are released under the \textit{Apache License 2.0}~\footnote{https://www.apache.org/licenses/LICENSE-2.0} where permissible and under \textit{AGPL-3.0 License}~\footnote{https://www.gnu.org/licenses/agpl-3.0.en.html} where mandated. These are described in Appendix~\ref{app:release}.

Models trained on the non-anonymized datasets will not be released for privacy reasons. Similarly, we will not be releasing the non-anonymized datasets publicly for privacy reasons.
\section{Conclusion}
This work introduces UVH-26, a large-scale, domain-specific dataset of annotated traffic-camera images from Indian urban environments, along with benchmark models fine-tuned on this data. By combining CCTV imagery, a structured crowdsourcing workflow, and consensus-based annotation strategies, UVH-26 addresses critical limitations of existing global datasets in representing heterogeneous, high-density traffic conditions from developing countries like India. Empirical evaluations demonstrate that models trained on UVH-26 achieve up to $31.5\%$ improvement in mAP over COCO-pretrained baselines, underscoring the importance of contextually aligned data for robust vehicle detection under surveillance viewpoints. The public release of both the dataset and fine-tuned models under permissive licenses provides a foundational resource for advancing intelligent transportation systems and computer vision research in emerging economies.
\section{Acknowledgments}
We thank the Bengaluru Traffic Police (BTP) and the Bengaluru Police for providing access to the Safe City camera data from which the image datasets used for this release were derived.
We thank Capital One for sponsoring the prizes for the \UVH competition. We thank IISc's AI and Robotics Technology Park (ARTPARK) and Centre for infrastructure, Sustainable Transportation and Urban Planning (CiSTUP) for funding the annotation and model training efforts, and Kotak IISc AI-ML Centre (KIAC) for providing the GPU resources required to train the models. 
We acknowledge the outreach support provided by the ACM India Council and IEEE India Council to help encourage chapter volunteers to participate in \UVH. 
Lastly, we thank the AI Centers of Excellence (AI COE) initiative of the Ministry of Education, their Apex Committee members, and the AIRAWAT Research Foundation who helped catalyze some of the early efforts.

\bibliographystyle{ieeetr}
\bibliography{references.bib}

\clearpage
\appendix

\section{Appendix: Release Notes}
\label{app:release}
The datasets and models are posted on Huggingface under \url{https://huggingface.co/iisc-aim/}. The datasets are under \url{https://huggingface.co/datasets/iisc-aim/UVH-26} while models are under \url{https://huggingface.co/iisc-aim/UVH-26}.  The \texttt{v1.0} branch in these repositories have the version discussed in this report.

\subsection{Datasets}
The folder structure on Huggingface for the datasets present under \url{https://huggingface.co/datasets/iisc-aim/UVH-26/tree/v1.0} is as follows:
\begin{itemize}
    \item \texttt{UVH-26-Train/}: This folder has the 80\% of \uvhw data used for training.
    \begin{itemize}
        \item \texttt{images/}: This folder contains the \uvhw training images organized, into sub-folder such as \texttt{000/, 001/, ...} for convenience.
        \item \texttt{images/000/*}: Actual training images with filenames \texttt{1.png, 2.png, ...} that are unique across all training images.
        \item \texttt{images/001/*}: ...
        \item \texttt{UVH-26-MV-Train.json}: Majority Voting Consensus Annotations for the training images, provided in COCO JSON format.
        \item \texttt{UVH-26-ST-Train.json}: STAPLE Consensus Annotations for the training images, provided in COCO JSON format.
    \end{itemize}
    \item \texttt{UVH-26-Val/}: This folder has the 20\% of \uvhw data used for validation.
    \begin{itemize}
        \item \texttt{images/}: This folder contains the \uvhw validation images organized, into sub-folder such as \texttt{000/, 001/, ...} for convenience.
        \item \texttt{images/002/*}: ...
        \item \texttt{UVH-26-MV-Val.json}: Majority Voting Consensus Annotations for the validation images, provided in COCO JSON format.
        \item \texttt{UVH-26-ST-Val.json}: STAPLE Consensus Annotations for the validation images, provided in COCO JSON format.
    \end{itemize}
    \item \texttt{LICENSE}: License file mentioning the \textit{Creative Commons Attribution 4.0 International License~\footnote{https://creativecommons.org/licenses/by/4.0/}} under which the data is being released. 
\end{itemize}

\subsection{Models}

The folder structure on Huggingface for the models present under \url{https://huggingface.co/iisc-aim/UVH-26/tree/v1.0} is as follows:
\begin{itemize}
    \item \texttt{uvh\_classes.txt}: Text file with the list of \uvh object classes used for training, with one class per line.
    \item \texttt{configs/}: Configuration files defining hyperparameters and architecture details used during model training.
    \begin{itemize}
        \item \texttt{yolo11\_x.yaml}, \texttt{yolo11\_s.yaml}, \texttt{rtdetr\_x.yaml}, \texttt{rtdetr\_s.yaml}, \texttt{damo\_yolo\_x.yaml}, \texttt{damo\_yolo\_s.yaml}: Configuration files used when training the six models.
    \end{itemize}
    \item \texttt{weights/}: Directory containing trained model weights organized by model family, size, and consensus dataset.
    \begin{itemize}
        \item \texttt{YOLOv11-X/}: Weights for the X-sized YOLOv11 model variant.
        \begin{itemize}
            \item \texttt{UVH-26-ST/}: Model trained on the \uvhst dataset.
            \begin{itemize}
                \item \texttt{UVH-26-ST-YOLOv11-X.pt}: Trainedweights in PyTorch format.
            \end{itemize}
            \item \texttt{UVH-26-MV/}: Model trained on the  dataset.
            \begin{itemize}
                \item \texttt{UVH-26-MV-YOLOv11-X.pt}: Trained weights in PyTorch format.
            \end{itemize}
        \end{itemize}
        \item \texttt{YOLOv11-S/}: Weights for the S-sized YOLOv11 model variant.
        \begin{itemize}
            \item \texttt{UVH-26-ST/}: Model trained on the \uvhst consensus dataset.
            \begin{itemize}
                \item \texttt{UVH-26-ST-YOLOv11-S.pt}: Trained weights in PyTorch format.
            \end{itemize}
            \item \texttt{UVH-26-MV/}: Model trained on the \uvhmv dataset.
            \begin{itemize}
                \item \texttt{UVH-26-MV-YOLOv11-S.pt}: Trained weights in PyTorch format.
            \end{itemize}
        \end{itemize}
        \item \texttt{RT-DETR-X/}: Weights for the X-sized RT-DETR model variant.
        \begin{itemize}
            \item \texttt{UVH-26-ST/}: Model trained on the \uvhst dataset.
            \begin{itemize}
                \item \texttt{UVH-26-ST-RT-DETR-X.pt}: Trained weights in PyTorch format.
            \end{itemize}
            \item \texttt{UVH-26-MV/}: Model trained on the \uvhmv dataset.
            \begin{itemize}
                \item \texttt{UVH-26-MV-RT-DETR-X.pt}: Trained weights in PyTorch format.
            \end{itemize}
        \end{itemize}

        \item \texttt{RT-DETR-S/}: Weights for the S-sized RT-DETR model variant.
        \begin{itemize}
            \item \texttt{UVH-26-ST/}: Model trained on the \uvhst dataset.
            \begin{itemize}
                \item \texttt{UVH-26-ST-RT-DETR-S.pt}: Trained weights in PyTorch format.
            \end{itemize}
            \item \texttt{UVH-26-MV/}: Model trained on the \uvhmv dataset.
            \begin{itemize}
                \item \texttt{UVH-26-MV-RT-DETR-S.pt}: Trained weights in PyTorch format.
            \end{itemize}
            
        \end{itemize}
        \item \texttt{DAMO-YOLO-X/}: Weights for the X-sized DAMO-YOLO model variant.
        \begin{itemize}
            \item \texttt{UVH-26-ST/}: Model trained on the \uvhst dataset.
            \begin{itemize}
                \item \texttt{UVH-26-ST-DAMO-YOLO-X.pt}: Trained weights in PyTorch format.
            \end{itemize}
            \item \texttt{UVH-26-MV/}: Model trained on the \uvhmv dataset.
            \begin{itemize}
                \item \texttt{UVH-26-MV-DAMO-YOLO-X.pt}: Trained weights in PyTorch format.
            \end{itemize}
        \end{itemize}

        \item \texttt{DAMO-YOLO-S/}: Weights for the S-sized DAMO-YOLO model variant.
        \begin{itemize}
            \item \texttt{UVH-26-ST/}: Model trained on the STAPLE consensus dataset.
            \begin{itemize}
                \item \texttt{UVH-26-ST-DAMO-YOLO-S.pt}: Trained weights in PyTorch format.
            \end{itemize}
            \item \texttt{UVH-26-MV/}: Model trained on the \uvhmv dataset.
            \begin{itemize}
                \item \texttt{UVH-26-MV-DAMO-YOLO-S.pt}: Trained weights in PyTorch format.
            \end{itemize}
        \end{itemize}
    \end{itemize}

    \item \texttt{usage/}: Folder for example scripts.
    \begin{itemize}
        \item \texttt{inference.py}: Script demonstrating how to load the model and perform inference on input images.
    \end{itemize}
    \item \texttt{LICENSE}: Specifies the usage terms for the released models and code, which follow the original licenses provided by their respective authors. Accordingly, the DAMO-YOLO and RT-DETR models are distributed under the \textit{Apache License 2.0}~\footnote{\url{https://www.apache.org/licenses/LICENSE-2.0}}, while the YOLO models are released under \textit{AGPL-3.0 License}~\footnote{\url{https://www.gnu.org/licenses/agpl-3.0.en.html}}, consistent with the terms defined in the source implementations. 
\end{itemize}

\clearpage
\section{Appendix: Dataset Details}
\label{add:dataset}

\subsection{\uvh Vehicle Classes}

\begin{table}[h]
\centering
\small
\caption{Vehicle Classes and Descriptions}
\label{tab:vehicle_category_description}
\begin{tabular}{L{4cm}|p{11cm}}
\hline
\textbf{Class} & \textbf{Description} \\
\hline
\hline
\textbf{Cycle} & Non-motorized, manually pedalled vehicles including geared, non-geared, women's, and children's cycles. Bounding boxes include both the vehicle and rider. \\
\hline
\textbf{2-Wheeler} & Motorbikes and scooters for single or double riders. Bounding boxes include both vehicle and rider. \\
\hline
\textbf{Bus} & Large passenger vehicles used for public or private transport, including office shuttles and intercity buses. \\
\hline
\textbf{M. Bus} & Shorter, compact buses with fewer seats. Larger than a Tempo Traveller, often featuring a flat front. \\
\hline
\textbf{Truck} & Heavy goods carriers with a front cabin and a rear cargo compartment. \\
\hline
\textbf{LCV (Light Commercial Vehicle)} & Lightweight goods carriers used for short to medium distance transport. \\
\hline
\textbf{T. Traveller} & Medium-sized passenger vans with tall roofs and side windows. Larger than vans but smaller than minibuses, with a protruding front. \\
\hline
\textbf{Van} & Medium-sized vehicles for transporting goods or people, typically with a flat front and sliding side doors. Smaller than Tempo Travellers. \\
\hline
\textbf{Sedan} & Passenger cars with a low-slung design and a separate protruding rear boot (“dickey”). \\
\hline
\textbf{Hatchback} & Small passenger cars without a protruding rear boot (“dickey”). \\
\hline
\textbf{SUV (Sport Utility Vehicle)} & Car-like vehicles with high ground clearance, a sturdy body, and no protruding boot. \\
\hline
\textbf{MUV (Multi-Utility Vehicle)} & Large vehicles with three seating rows, combining passenger and cargo functionality. \\
\hline
\textbf{3-Wheeler (Auto-rickshaw)} & Compact vehicles with one front wheel and two rear wheels, featuring a covered passenger cabin. \\
\hline
\textbf{Other} & Vehicles not covered in other classes, including agricultural, specialized, or unconventional designs. \\
\hline
\end{tabular}
\end{table}

\subsection{Camera Locations}
\label{app:cameras}
\begin{figure}[h]
    \centering
    \includegraphics[width=0.65\textwidth]{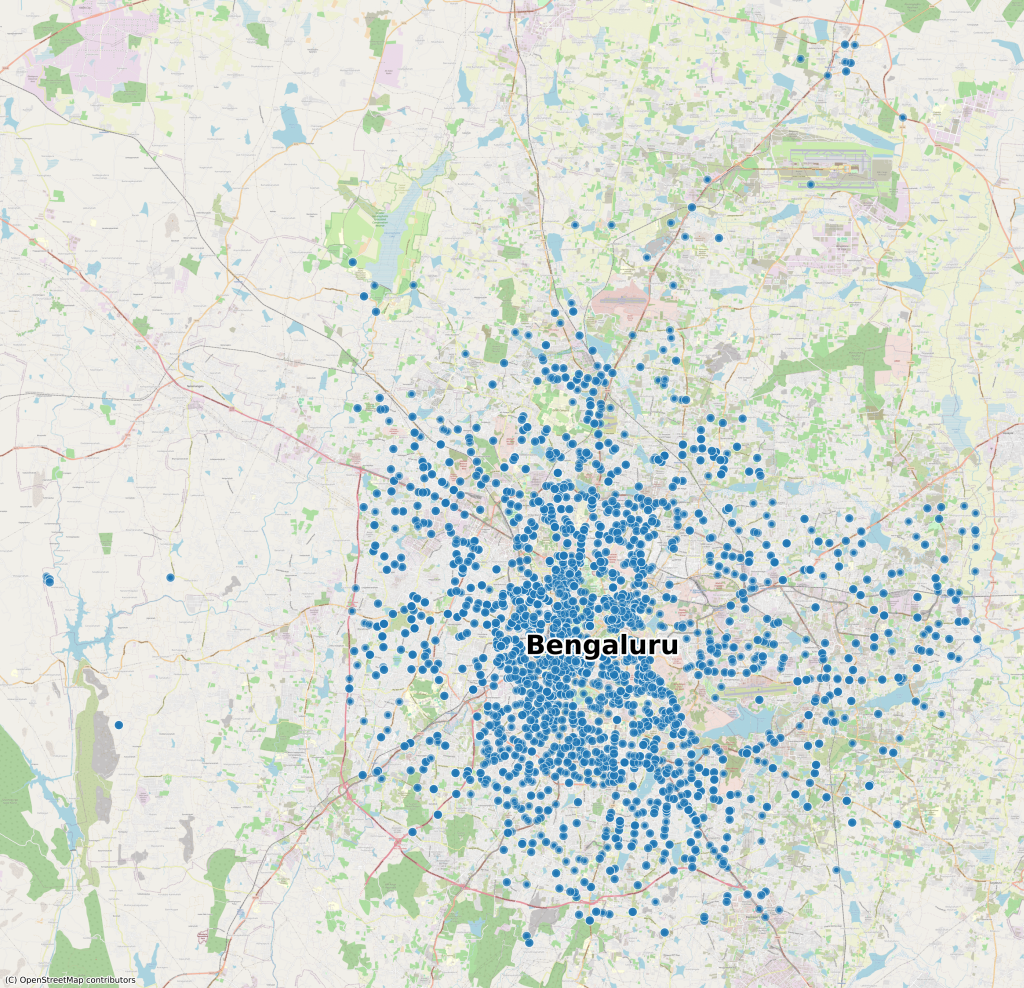}
    \caption{Location of Safe City cameras in Bengaluru from which images for \uvhw were sourced.}
    \label{fig:camera-map}
\end{figure}

Figure~\ref{fig:camera-map} shows the location of the 2800 Safe City cameras in Bengaluru from which images for \uvhw were sourced.

\subsection{Difficulty and Disagreement Scores}
\label{app:difficulty}

\paragraph{Notation.} Let $M$ be the number of detectors/models (or annotators) used for comparison and $C$ the number of classes. For a given image, let $c_{m,i}$ denote the count of bounding boxes of class $i$ predicted by model $m$ ($m\in\{1,\dots,M\}$, $i\in\{1,\dots,C\}$). Define $B_m=\sum_{i=1}^C c_{m,i}$ as the total bounding-box count produced by model $m$. We use $i$ as an image index where required; when ambiguity is possible we write $c_{m,i}^{(img)}$ or $D_i$ for the image-level disagreement score of image $i$.

\subsubsection{Disagreement Score}
The disagreement score captures four complementary aspects of inter-model variability.
\begin{itemize}
  \item \textbf{Per-class count disagreement.} For each class $i$ compute the standard deviation of counts across models:
  \begin{equation}
    \sigma(c_i)=\sqrt{\frac{1}{M}\sum_{m=1}^M \big(c_{m,i}-\overline{c_i}\big)^2},
    \qquad
    \overline{c_i}=\frac{1}{M}\sum_{m=1}^M c_{m,i}.
  \end{equation}
  Summing across classes yields the per-image class-count disagreement:
  \begin{equation}
    N_{dci}=\sum_{i=1}^C \sigma(c_i).
  \end{equation}
  This term measures how much the models disagree on counts for each class (e.g., some models see three three-wheelers while others see one).
    
  \item \textbf{Maximum pairwise class-count disagreements.} For each class $i$ count how many model pairs disagree in their class counts:
  \begin{equation}
    D_i = \sum_{m=1}^{M-1}\sum_{n=m+1}^{M} \mathbb{I}\big(c_{m,i} \neq c_{n,i}\big).
  \end{equation}
  Then take the worst (maximum) across classes:
  \begin{equation}
    M_{mdi} = \max_{i\in\{1,\dots,C\}} D_i.
  \end{equation}
  $M_{mdi}$ highlights the single class with the largest pairwise disagreement and emphasizes hard, contested categories.
\end{itemize}

We combine the main components into a compact per-image disagreement score:
\begin{equation}
  D_i = N_{dci} + M_{mdi},
\end{equation}
which balances aggregate count variance with the worst-case per-class pairwise disagreement. To make scores comparable across the dataset we normalize:
\begin{equation}
  D_i^{\mathrm{norm}} = \frac{D_i - D_{\min}}{D_{\max} - D_{\min}} \times 100,
\end{equation}
where $D_{\min}$ and $D_{\max}$ are the observed minimum and maximum $D_i$ values. All component quantities used in selection (e.g., $V_{uci},\,V_{nbi},\,N_{dci},\,M_{mdi}$) are retained for analysis and can be inspected individually when diagnosing why a particular image is contentious.

\subsubsection{Difficulty Score}
While disagreement measures \emph{inter-model uncertainty} (useful to prioritize images where detectors disagree), it does not by itself quantify visual complexity. To ensure annotator workload was balanced and to construct zone-wise difficulty progression in the gamified challenge, we computed a complementary image-level \emph{difficulty score} that captures intrinsic visual factors (object count, scale, density, overlap) together with model disagreement.

\paragraph{Definitions.} For a given image of resolution $H\times W$ with $N_{\mathrm{bboxes}}$ detected or annotated boxes $B_1,\dots,B_{N_{\mathrm{bboxes}}}$ (each box $B_j$ has width $w_j$ and height $h_j$), we compute the following normalized components:
\begin{itemize}
  \item \textbf{Bounding-box count:} 
  \begin{equation}
    M_{\text{bbox\_count}} = N_{\mathrm{bboxes}}
  \end{equation}
  normalized by a dataset maximum $M_{\text{bb\_max}}$:
  \begin{equation}
    \tilde{C} = \frac{M_{\text{bbox\_count}}}{M_{\text{bb\_max}}}\in[0,1].
  \end{equation}

  \item \textbf{Average box size:} the mean relative area
  \begin{equation}
    M_{\text{bbox\_size}} = \frac{1}{HW}\cdot\frac{1}{N_{\mathrm{bboxes}}}\sum_{j=1}^{N_{\mathrm{bboxes}}} w_j h_j,
  \end{equation}
  and we use its complement 
  \begin{equation}
    (1 - M_{\text{bbox\_size}})\in[0,1]
  \end{equation}
  so that smaller average objects increase difficulty.

  \item \textbf{Bounding-box density:} total box area fraction
  \begin{equation}
    M_{\text{bbox\_density}} = \frac{1}{HW}\sum_{j=1}^{N_{\mathrm{bboxes}}} w_j h_j,
  \end{equation}
  clipped or normalized into $[0,1]$ (we use $\min(1, M_{\text{bbox\_density}})$).

  \item \textbf{Class diversity:} 
  \begin{equation}
    M_{\text{class\_count}} = |\{\text{unique classes in image}\}|
  \end{equation}
  normalized by the maximum number of classes $M_{\text{max\_classes}}$:
  \begin{equation}
    \tilde{K} = \frac{M_{\text{class\_count}}}{M_{\text{max\_classes}}}\in[0,1].
  \end{equation}

  \item \textbf{Average IoU overlap:} for each unordered box pair $(B_p,B_q)$ define
  \begin{equation}
    \mathrm{IoU}(B_p,B_q)=\frac{|B_p\cap B_q|}{|B_p\cup B_q|},
  \end{equation}
  and the mean pairwise overlap
  \begin{equation}
    M_{\mathrm{iou\_overlap}} = \frac{1}{N_{\mathrm{pairs}}}\sum_{p<q}\mathrm{IoU}(B_p,B_q),
    \quad N_{\mathrm{pairs}}=\binom{N_{\mathrm{bboxes}}}{2}.
  \end{equation}
  High $M_{\mathrm{iou\_overlap}}$ indicates occlusion and crowding.
  
  \item \textbf{Model disagreement (normalized):} we reuse the disagreement score above and scale it to $[0,1]$:
  \begin{equation}
    \widetilde{D}_i = \frac{D_i^{\mathrm{norm}}}{100}\in[0,1].
  \end{equation}
\end{itemize}

\paragraph{Composite difficulty score.} The image difficulty $\Delta_i$ is a weighted sum of normalized components:
\begin{equation}
  \Delta_i = \tilde{C}\;+\; (1 - M_{\text{bbox\_size}})\;+\; \widetilde{D}_i\;+\; M_{\mathrm{iou\_overlap}},
\end{equation}
here all components are pre-normalized to $[0,1]$. Optionally, one can include $\tilde{K}$ (class diversity) or $M_{\text{bbox\_density}}$ as additional terms if finer control is required. Finally, as with disagreement, $\Delta_i$ can be rescaled to $[0,100]$ for presentation.

\subsection{Average Precision Metrics Used}
\label{app:map}

\begin{enumerate}
  \item mAP(50:95): the main benchmark metric, defined as the mean of AP values at IoU thresholds from \(0.50\) to \(0.95\) in steps of \(0.05\).  
  For a predicted box \(B\) and ground truth \(G\), the Intersection-over-Union (IoU) is
  \begin{equation}
    \mathrm{IoU}(B,G)=\frac{|B\cap G|}{|B\cup G|}.
  \end{equation}
  For class \(c\) at a fixed threshold \(t\), the average precision is
  \begin{equation}
    \mathrm{AP}_{c}(t)=\int_{0}^{1} p^{\mathrm{interp}}_{c,t}(r)\,dr,
  \end{equation}
  where \(p^{\mathrm{interp}}_{c,t}(r)\) is the \emph{interpolated precision} function defined as
  \begin{equation}
    p^{\mathrm{interp}}_{c,t}(r)=\max_{\tilde{r}\ge r} p_{c,t}(\tilde{r}),
  \end{equation}
  ensuring a non-increasing precision–recall curve as used in COCO evaluations.  
  The mean over all classes is
  \begin{equation}
    \mathrm{mAP}(t)=\frac{1}{C}\sum_{c=1}^{C}\mathrm{AP}_{c}(t).
  \end{equation}
  Combining across thresholds gives
  \begin{equation}
    \mathrm{mAP}(50{:}95)=\frac{1}{10}\sum_{j=0}^{9}\mathrm{mAP}\!\bigl(0.50+0.05\,j\bigr).
  \end{equation}
  This provides a comprehensive measure of both detection accuracy and localization robustness.

  \item mAP(75): AP computed at a stricter IoU threshold of \(0.75\), which emphasizes precise localization quality:
  \begin{equation}
    \mathrm{mAP}(75)=\frac{1}{C}\sum_{c=1}^{C}\mathrm{AP}_{c}(0.75).
  \end{equation}

  \item mAP(50): AP computed at a lenient IoU threshold of \(0.50\), reflecting the model’s capacity for coarse but correct detections:
  \begin{equation}
    \mathrm{mAP}(50)=\frac{1}{C}\sum_{c=1}^{C}\mathrm{AP}_{c}(0.50).
  \end{equation}
\end{enumerate}

\subsection{Hyperparameters used for Model Training}
\label{app:hyperparam}

\begin{table}[h]
    \centering
    \caption{Training hyperparameters and architectural settings for training models in \uvhmv.}
    \scriptsize
    \setlength{\tabcolsep}{2pt}
    \renewcommand{\arraystretch}{1.08}
    \begin{tabular}{p{2.25cm}|p{2.25cm}p{2.25cm}p{2.25cm}p{2.25cm}p{2.25cm}p{2.25cm}}
        \hline
        \textbf{Setting}
        & \textbf{DAMO-YOLO-T}
        & \textbf{DAMO-YOLO-L}
        & \textbf{YOLOv11-S}
        & \textbf{YOLOv11-X}
        & \textbf{RT-DETRv2-S}
        & \textbf{RT-DETRv2-X}\\
        \hline\hline

        Batch/Best Epoch
        & 16 / 47
        & 16 / 18
        & 16 / 30
        & 16 / 13
        & 16 / 66
        & 16 / 68\\

        LR / Opt.
        & 0.01/64 SGD
        & 0.01/64 SGD
        & auto/AdamW
        & auto/AdamW
        & 1e-4/AdamW
        & 1e-4/AdamW \\

        Decay / Mom.
        & 5e-4/0.9
        & 5e-4/0.9
        & 5e-4/(0.9, 0.999)
        & 5e-4/(0.9, 0.999)
        & 1e-4/(0.9, 0.999)
        & 1e-4/(0.9, 0.999) \\

        LR Policy
        & Const.
        & Const.
        & Cosine
        & Cosine
        & Linear
        & Linear \\

        Warmup / No-Aug
        & 5/16
        & 5/16
        & patience=150
        & patience=150
        & --, 70
        & --, 70 \\

        Augmentation
        & mixup=0.15, shear=$2^\circ$, deg=10
        & mixup=0.15, shear=$2^\circ$, deg=10
        & no mosaic/mixup
        & no mosaic/mixup
        & flip, color distort, zoom, IoU crop (till 70e)
        & flip, colordistort, zoom, IoU crop (till 70e) \\

        Backbone
        & TinyNAS-L20 + Giraffe NeckV2
        & TinyNAS-L45+ Giraffe NeckV2
        & YOLOv11s + CSP
        & YOLOv11x + CSP
        & PResNet-18 + HybridEnc
        & PResNet-101 + HybridEnc\\
        \hline
    \end{tabular}
    \label{tab:training_hparams}
\end{table}

\clearpage

\clearpage
\section{Comparison of Models Fine-tuned on \uvhmv Dataset (Anonymized)}
\label{app:anon_datasets}

This section reports additional evaluation details for the six models trained on the \uvhmv dataset and annotations, in comparison with the baseline models trained on COCO dataset.

\definecolor{damo_color}{HTML}{c2c3df}
\definecolor{best_color}{HTML}{fcab8f}
\definecolor{best_color}{HTML}{aaffaa}
\definecolor{best2_color}{HTML}{aaddff}
\definecolor{yolov8_color}{HTML}{a9dca3}
\definecolor{yolov11_color}{HTML}{a6cee4}

\begin{table}[htbp]
\centering
\caption{\textit{Per-class AP metrics} for the three classes common to COCO and \uvh (Car, Bus, Truck) on the held out test set\tablefootnote{Blue cells indicate the best score within a training data group and metric column, across models. Green cells indicate the best score within a metric column, across training data groups and models.}.
 }

\scriptsize
\begin{tabular}{lc rrr rrr rrr}
\toprule
& & \multicolumn{3}{c}{\bf Car AP} & \multicolumn{3}{c}{\bf Bus AP} & \multicolumn{3}{c}{\bf Truck AP} \\
\cmidrule(lr){3-5} \cmidrule(lr){6-8} \cmidrule(lr){9-11}
\bf Model & \bf Training Data & \em 50:95 & \em 75 & \em 50 & \em 50:95 & \em 75 & \em 50 & \em 50:95 & \em 75 & \em 50 \\
\hline\hline

YOLO11S      & Majority Voting & 0.5701 & 0.6757 & 0.7225 & 0.2579 & 0.2793 & 0.3677 & 0.4600 & 0.5428 & 0.6421 \\
YOLO11X      & Majority Voting & 0.5818 & 0.6866 & 0.7258 & 0.3538 & 0.3730 & 0.4698 & 0.4877 & 0.5597 & 0.6763 \\
RT-DETR-S    & Majority Voting & 0.7082 & 0.8300 & 0.8876 & 0.5969 & 0.6605 & 0.7891 & 0.6271 & 0.7352 & 0.8389 \\
RT-DETR-X    & Majority Voting & \cellcolor{best2_color}\textbf{0.7291} & \cellcolor{best2_color}\textbf{0.8523} & \cellcolor{best2_color}\textbf{0.9081} & \cellcolor{best2_color}\textbf{0.6168} & \cellcolor{best2_color}\textbf{0.6801} & \cellcolor{best2_color}\textbf{0.8365} & \cellcolor{best2_color}\textbf{0.6658} & \cellcolor{best2_color}\textbf{0.7921} & \cellcolor{best2_color}\textbf{0.8927} \\
DAMO-YOLO-T  & Majority Voting & 0.6171 & 0.7399 & 0.8289 & 0.5295 & 0.5958 & 0.7535 & 0.5175 & 0.5767 & 0.7965 \\
DAMO-YOLO-L  & Majority Voting & 0.6465 & 0.7771 & 0.8478 & 0.5624 & 0.6515 & 0.7962 & 0.5962 & 0.7276 & 0.8537 \\
\midrule
\addlinespace[0.8em]

YOLO11S      & COCO weights & 0.5402 & 0.6171 & 0.7285 & 0.2261 & 0.2579 & 0.2672 & 0.1502 & 0.1716 & 0.2026 \\
YOLO11X      & COCO weights & 0.5577 & 0.6387 & 0.7132 & \cellcolor{best_color}\textbf{0.3189} & \cellcolor{best_color}\textbf{0.3683} & 0.3739 & 0.2937 & 0.3420 & 0.3993 \\
RT-DETR-S    & COCO weights & 0.6265 & 0.7112 & 0.8335 & 0.1967 & 0.2214 & 0.2464 & 0.1664 & 0.1653 & 0.2472 \\
RT-DETR-X    & COCO weights & \cellcolor{best_color}\textbf{0.6490} & \cellcolor{best_color}\textbf{0.7499} & 0.8582 & 0.2509 & 0.2748 & 0.3093 & \cellcolor{best_color}\textbf{0.3098} & \cellcolor{best_color}\textbf{0.3424} & \cellcolor{best_color}\textbf{0.4579} \\
DAMO-YOLO-T  & COCO weights & 0.5647 & 0.6340 & 0.8036 & 0.2055 & 0.2301 & 0.2599 & 0.0996 & 0.1033 & 0.1544 \\
DAMO-YOLO-L  & COCO weights & 0.6479 & 0.7370 & \cellcolor{best_color}\textbf{0.8594} & 0.3108 & 0.3573 & \cellcolor{best_color}\textbf{0.3855} & 0.2185 & 0.2533 & 0.3076 \\
\bottomrule
\end{tabular}
\label{tab:uvh_per_class_metrics_blurred}
\end{table}

\begin{table}[ht]
    \centering
    \caption{\textit{Overall mAP metrics} across all three classes common to COCO and \uvh (Car, Bus, Truck). Each mAP value is averaged over the per-class AP for the three classes\tablefootnote{Blue cells indicate the best score within a training data group and metric column, across models. Green cells indicate the best score within a metric column, across training data groups and models.}
     }
    \footnotesize
  
    \begin{tabular}{lcrrr}
        \toprule
        \bf Model & \bf Method & \bf mAP(50:95) & \bf mAP(75) & \bf mAP(50) \\
        \hline\hline

        YOLO11S     & Majority Voting & 0.4293 & 0.4993 & 0.5774 \\
        YOLO11X     & Majority Voting & 0.4744 & 0.5397 & 0.6240 \\
        RT-DETR-S   & Majority Voting & 0.6441 & 0.7419 & 0.8385 \\
        RT-DETR-X   & Majority Voting & \cellcolor{best2_color}\textbf{0.6706} & \cellcolor{best2_color}\textbf{0.7748} & \cellcolor{best2_color}\textbf{0.8791} \\
        DAMO-YOLO-T & Majority Voting & 0.5547 & 0.6375 & 0.7930 \\
        DAMO-YOLO-L & Majority Voting & 0.6017 & 0.7187 & 0.8326 \\

        \midrule
        \addlinespace[0.5em]

        YOLO11S     & COCO weights & 0.3055 & 0.3489 & 0.3994 \\
        YOLO11X     & COCO weights & 0.3901 & 0.4497 & 0.4955 \\
        RT-DETR-S   & COCO weights & 0.3299 & 0.3660 & 0.4424 \\
        RT-DETR-X   & COCO weights & \cellcolor{best_color}\textbf{0.4033} & \cellcolor{best_color}\textbf{0.4557} & \cellcolor{best_color}\textbf{0.5418} \\
        DAMO-YOLO-T & COCO weights & 0.2899 & 0.3225 & 0.4060 \\
        DAMO-YOLO-L & COCO weights & 0.3924 & 0.4492 & 0.5175 \\
        \bottomrule    
    \end{tabular}
    \label{tab:uvh_overall_metrics_blurred}
\end{table}

\begin{sidewaystable}[htbp]
\centering
\caption{Per-class AP for all \uvh classes for different models~\tablefootnote{Green color is best score and blue is second best score for a class (maxima in a row), for each AP metric.}.}
\scriptsize

\begin{tabular}{lccccccccccccccccccc}
\toprule
& \multicolumn{3}{c}{\bf YOLO11S} & \multicolumn{3}{c}{\bf YOLO11X} & \multicolumn{3}{c}{\bf RT-DETR-S} & \multicolumn{3}{c}{\bf RT-DETR-X} & \multicolumn{3}{c}{\bf DAMO-YOLO-T} & \multicolumn{3}{c}{\bf DAMO-YOLO-L} \\
\cmidrule(lr){2-4} \cmidrule(lr){5-7} \cmidrule(lr){8-10} \cmidrule(lr){11-13} \cmidrule(lr){14-16} \cmidrule(lr){17-19}
\textbf{Class} & 50 & 75 & 50:95 & 50 & 75 & 50:95 & 50 & 75 & 50:95 & 50 & 75 & 50:95 & 50 & 75 & 50:95 & 50 & 75 & 50:95 \\
\midrule

\bf Cycle &
0.2853 & 0.2489 & 0.2075 &
0.3511 & 0.2581 & 0.2407 &
\cellcolor{best2_color}\textbf{0.7130} & \cellcolor{best2_color}\textbf{0.4655} & \cellcolor{best2_color}\textbf{0.4384} &
\cellcolor{best_color}\textbf{0.7959} & \cellcolor{best_color}\textbf{0.5687} & \cellcolor{best_color}\textbf{0.5134} &
0.5626 & 0.3849 & 0.3496 &
0.6241 & 0.3894 & 0.3831 \\[0.45em]

\bf 2-Wheeler &
0.6950 & 0.4565 & 0.4249 &
0.7240 & 0.4951 & 0.4541 &
\cellcolor{best2_color}\textbf{0.8941} & \cellcolor{best2_color}\textbf{0.6244} & \cellcolor{best2_color}\textbf{0.5677} &
\cellcolor{best_color}\textbf{0.9036} & \cellcolor{best_color}\textbf{0.6526} & \cellcolor{best_color}\textbf{0.5831} &
0.8506 & 0.5265 & 0.4966 &
0.8759 & 0.5475 & 0.5135 \\[0.45em]

\bf Bus &
0.3115 & 0.2166 & 0.2118 &
0.3971 & 0.2999 & 0.2912 &
\cellcolor{best2_color}\textbf{0.6975} & \cellcolor{best2_color}\textbf{0.5602} & \cellcolor{best2_color}\textbf{0.5170} &
\cellcolor{best_color}\textbf{0.7239} & \cellcolor{best_color}\textbf{0.5661} & \cellcolor{best_color}\textbf{0.5211} &
0.6430 & 0.5061 & 0.4574 &
0.6831 & 0.5571 & 0.4859 \\[0.45em]

\bf M. Bus &
0.0297 & 0.0297 & 0.0178 &
0.0817 & 0.0817 & 0.0661 &
\cellcolor{best2_color}\textbf{0.1976} & \cellcolor{best2_color}\textbf{0.1921} & \cellcolor{best2_color}\textbf{0.1495} &
0.1941 & 0.1664 & 0.1412 &
\cellcolor{best_color}\textbf{0.2727} & \cellcolor{best_color}\textbf{0.2553} & \cellcolor{best_color}\textbf{0.1903} &
0.2072 & 0.1801 & 0.1397 \\[0.45em]

\bf Truck &
0.5220 & 0.5022 & 0.3861 &
0.5617 & 0.5090 & 0.4119 &
\cellcolor{best2_color}\textbf{0.7034} & \cellcolor{best2_color}\textbf{0.6299} & \cellcolor{best2_color}\textbf{0.5314} &
\cellcolor{best_color}\textbf{0.7702} & \cellcolor{best_color}\textbf{0.6975} & \cellcolor{best_color}\textbf{0.5913} &
0.6931 & 0.5001 & 0.4463 &
0.7362 & 0.6616 & 0.5182 \\[0.45em]

\bf LCV &
0.6165 & 0.5074 & 0.4392 &
0.6301 & 0.5153 & 0.4553 &
\cellcolor{best2_color}\textbf{0.8328} & \cellcolor{best2_color}\textbf{0.7203} & \cellcolor{best2_color}\textbf{0.6146} &
\cellcolor{best_color}\textbf{0.8717} & \cellcolor{best_color}\textbf{0.7635} & \cellcolor{best_color}\textbf{0.6398} &
0.7744 & 0.5534 & 0.4983 &
0.8237 & 0.6881 & 0.5716 \\[0.45em]

\bf T. Traveller &
0.8031 & 0.7735 & 0.7062 &
0.8145 & 0.8145 & 0.7668 &
\cellcolor{best2_color}\textbf{0.9611} & \cellcolor{best2_color}\textbf{0.9181} & \cellcolor{best2_color}\textbf{0.8711} &
\cellcolor{best_color}\textbf{0.9668} & \cellcolor{best_color}\textbf{0.9168} & \cellcolor{best_color}\textbf{0.8722} &
0.8950 & 0.8809 & 0.7849 &
0.9370 & 0.9024 & 0.8163 \\[0.45em]

\bf Van &
0.4846 & 0.4846 & 0.3913 &
0.3258 & 0.3258 & 0.2798 &
\cellcolor{best2_color}\textbf{0.8156} & \cellcolor{best2_color}\textbf{0.7940} & \cellcolor{best2_color}\textbf{0.6619} &
\cellcolor{best_color}\textbf{0.8244} & \cellcolor{best_color}\textbf{0.7908} & \cellcolor{best_color}\textbf{0.6666} &
0.7697 & 0.7224 & 0.5750 &
0.8468 & 0.8100 & 0.6503 \\[0.45em]

\bf Sedan &
0.4491 & 0.4232 & 0.3626 &
0.4471 & 0.4104 & 0.3537 &
\cellcolor{best2_color}\textbf{0.7236} & \cellcolor{best2_color}\textbf{0.6546} & \cellcolor{best2_color}\textbf{0.5613} &
\cellcolor{best_color}\textbf{0.7579} & \cellcolor{best_color}\textbf{0.6907} & \cellcolor{best_color}\textbf{0.5968} &
0.6976 & 0.5953 & 0.4989 &
0.7469 & 0.6586 & 0.5510 \\[0.45em]

\bf Hatchback &
0.4512 & 0.4300 & 0.3624 &
0.5084 & 0.4851 & 0.4126 &
\cellcolor{best2_color}\textbf{0.6226} & \cellcolor{best2_color}\textbf{0.5955} & \cellcolor{best2_color}\textbf{0.5121} &
\cellcolor{best_color}\textbf{0.6483} & \cellcolor{best_color}\textbf{0.6249} & \cellcolor{best_color}\textbf{0.5370} &
0.6289 & 0.5751 & 0.4798 &
0.6306 & 0.5855 & 0.4885 \\[0.45em]

\bf SUV &
0.2321 & 0.2217 & 0.1976 &
0.2531 & 0.2424 & 0.2206 &
\cellcolor{best2_color}\textbf{0.5933} & \cellcolor{best2_color}\textbf{0.5536} & \cellcolor{best2_color}\textbf{0.4806} &
\cellcolor{best_color}\textbf{0.6514} & \cellcolor{best_color}\textbf{0.6137} & \cellcolor{best_color}\textbf{0.5331} &
0.5315 & 0.4655 & 0.4035 &
0.6010 & 0.5571 & 0.4734 \\[0.45em]

\bf MUV &
0.1473 & 0.1402 & 0.1257 &
0.1689 & 0.1604 & 0.1421 &
\cellcolor{best2_color}\textbf{0.4866} & \cellcolor{best2_color}\textbf{0.4720} & \cellcolor{best2_color}\textbf{0.3865} &
\cellcolor{best_color}\textbf{0.5812} & \cellcolor{best_color}\textbf{0.5542} & \cellcolor{best_color}\textbf{0.4652} &
0.5382 & 0.5074 & 0.3990 &
0.5648 & 0.5271 & 0.4368 \\[0.45em]

\bf 3-Wheeler &
0.8042 & 0.6588 & 0.5747 &
0.8144 & 0.6765 & 0.5857 &
\cellcolor{best2_color}\textbf{0.9331} & \cellcolor{best2_color}\textbf{0.7811} & \cellcolor{best2_color}\textbf{0.6779} &
\cellcolor{best_color}\textbf{0.9332} & \cellcolor{best_color}\textbf{0.7973} & \cellcolor{best_color}\textbf{0.6866} &
0.9083 & 0.7036 & 0.6102 &
0.9269 & 0.7361 & 0.6352 \\
\bottomrule
\end{tabular}
\label{tab:hackathon_per_class_merged_blurred}
\end{sidewaystable}

\begin{table}[ht]
    \centering
    \caption{
    \textit{Overall mAP metrics} across all \uvh classes per model. Each mAP value is averaged over the per-class AP for all \uvh classes~\tablefootnote{Green color is best score and blue is second best score for column.}.}
    \footnotesize
    \begin{tabular}{lrrr}
        \toprule
        \bf Model & \bf mAP(50) & \bf mAP(50:95) & \bf mAP(75) \\
        \hline\hline
        YOLO11S      & 0.4486 & 0.3391 & 0.3918 \\
        YOLO11X      & 0.4675 & 0.3600 & 0.4057 \\
        RT-DETR-S    & 0.7057 & \cellcolor{best2_color}\textbf{0.5362} & \cellcolor{best2_color}\textbf{0.6124} \\
        RT-DETR-X    & \cellcolor{best_color}\textbf{0.7402} & \cellcolor{best_color}\textbf{0.5652} & \cellcolor{best_color}\textbf{0.6464} \\
        DAMO-YOLO-T  & 0.6743 & 0.4761 & 0.5520 \\
        DAMO-YOLO-L  & \cellcolor{best2_color}\textbf{0.7080} & 0.5126 & 0.6000 \\
        \bottomrule
    \end{tabular}
    \label{tab:hackathon_overall_map_blurred}
\end{table}

\clearpage
\section{Comparison of Models on \textit{Non-anonymized} \uvhw Dataset}
\label{app:anon_datasets:brief}
While we are not placing the non-anonymized \uvhw dataset in the public domain, we did fine-tune the baseline models using the non-anonymized \uvhw dataset to compare the difference in accuracy between models fine-tuned with and without anonymizations. In this section, we report results for the mode trained on the non-anonymized \uvhw dataset. Here, we include models fined-tuned using both \textit{Majority Voting} and \textit{STAPLE} consensus over the non-anonymized datasets. As mentioned, for privacy reasons, \textbf{we are not releasing} the non-anonymized datasets or models in the public domain, and these results are reported only for academic benefit.

\begin{figure}[t]
    \centering
    \subfloat[COCO and \uvhw models on common classes.]{\includegraphics[width=0.45\textwidth]{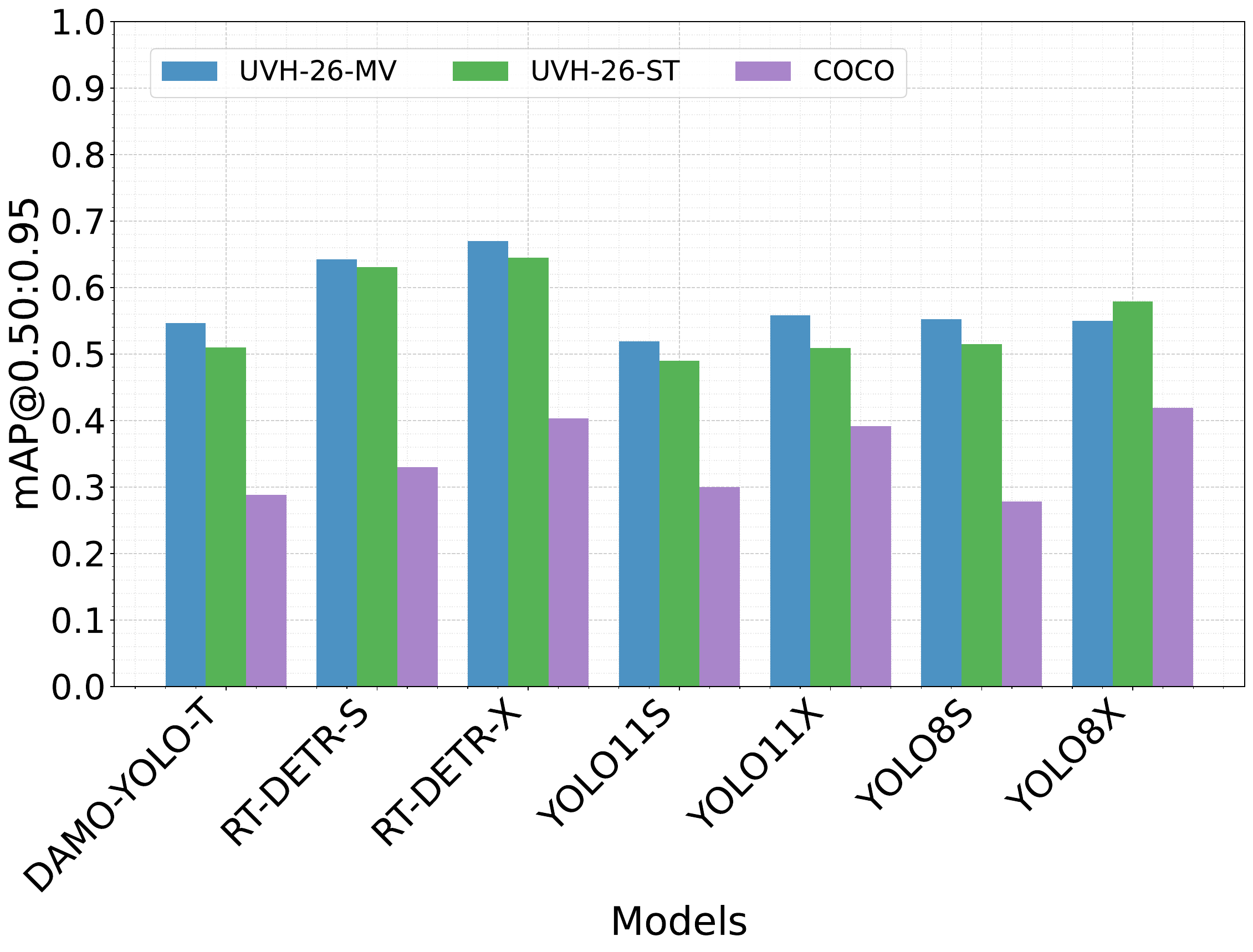}\label{fig:app-hackathon_ap_50_95_performance}}
        \hfill
    \subfloat[\uvhmv and \uvhst models on all \uvh classes.]{
    \includegraphics[width=0.45\textwidth]{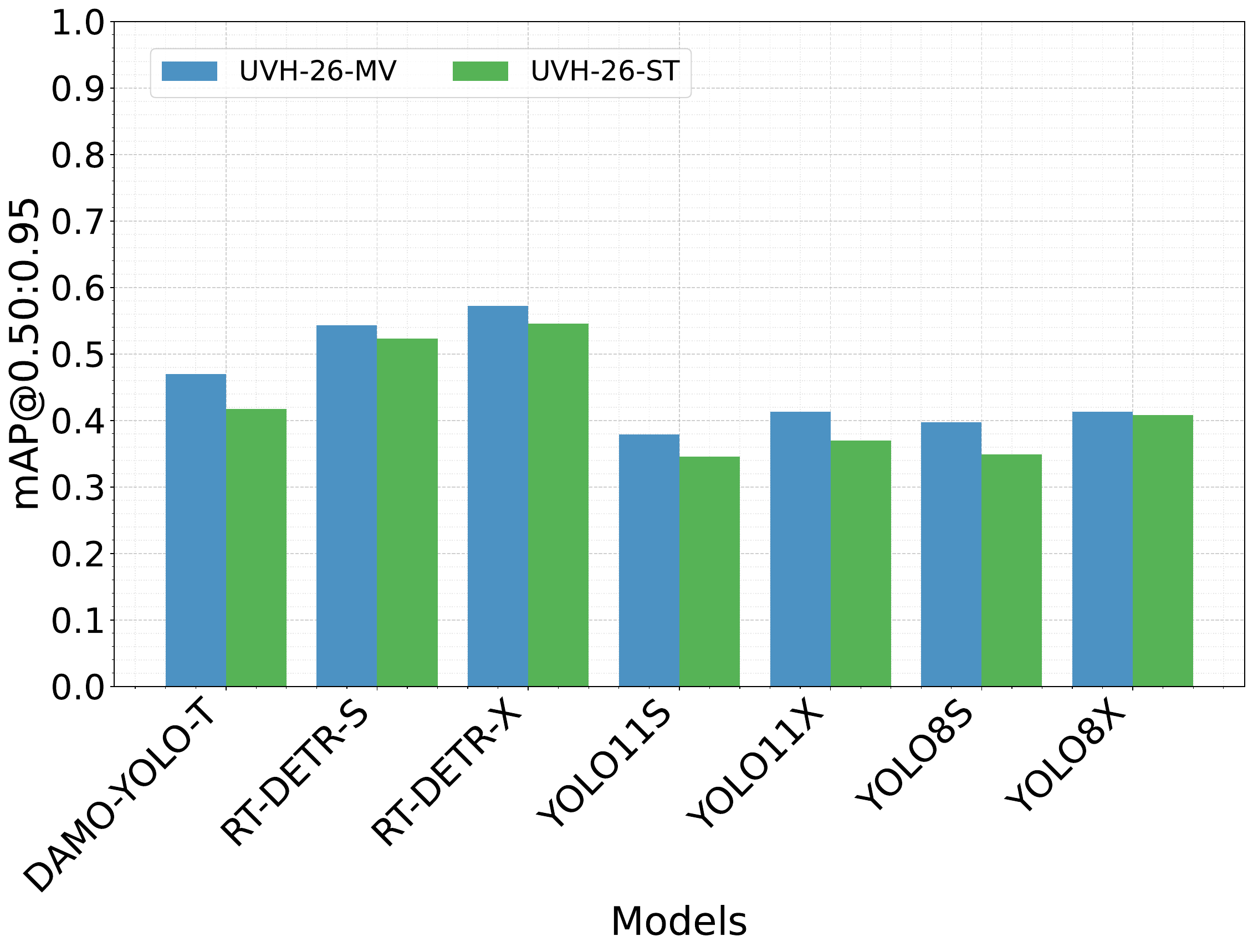}
    \label{fig:app-hackathon_ap_50_95_performance_2}}
    \caption{Overall performance (mAP@50:95)
    of all model architectures on the \textbf{non-anonymized} \uvhw datasets. Per-class APs are averaged across all relevant classes.}
    \label{fig:app-difficulty}
\end{figure}

The overall and per-class metrics are provided in Table~\ref{tab:hackathon_overall_map}, Table~\ref{tab:hackathon_per_class_merged}, and Figures~\ref{fig:app-hackathon_ap_50_95_performance}-\ref{fig:app-hackathon_ap_50_95_performance_2}.
Across architectures, models trained on MV show small but consistent gains over STAPLE on $mAP(50:95)$ and, more noticeably, at higher IoU thresholds. This pattern suggests that the estimated ground truth from Majority Voting provides better supervision for box labels as compared to STAPLE. Architecture effects are stable across consensus choices. Transformer-based RT-DETR variants rank highest on $mAP(50:95)$, indicating better localization under dense scenes and wide fields of view of top-down CCTV. Convolutional families (YOLO and DAMO-YOLO) have comparable results at $AP(50)$ but they show a larger drop at $AP(75)$ relative to RT-DETR, pointing to finer localization advantages from global attention.

\begin{figure}[h]
    \centering
    \includegraphics[width=1\textwidth]{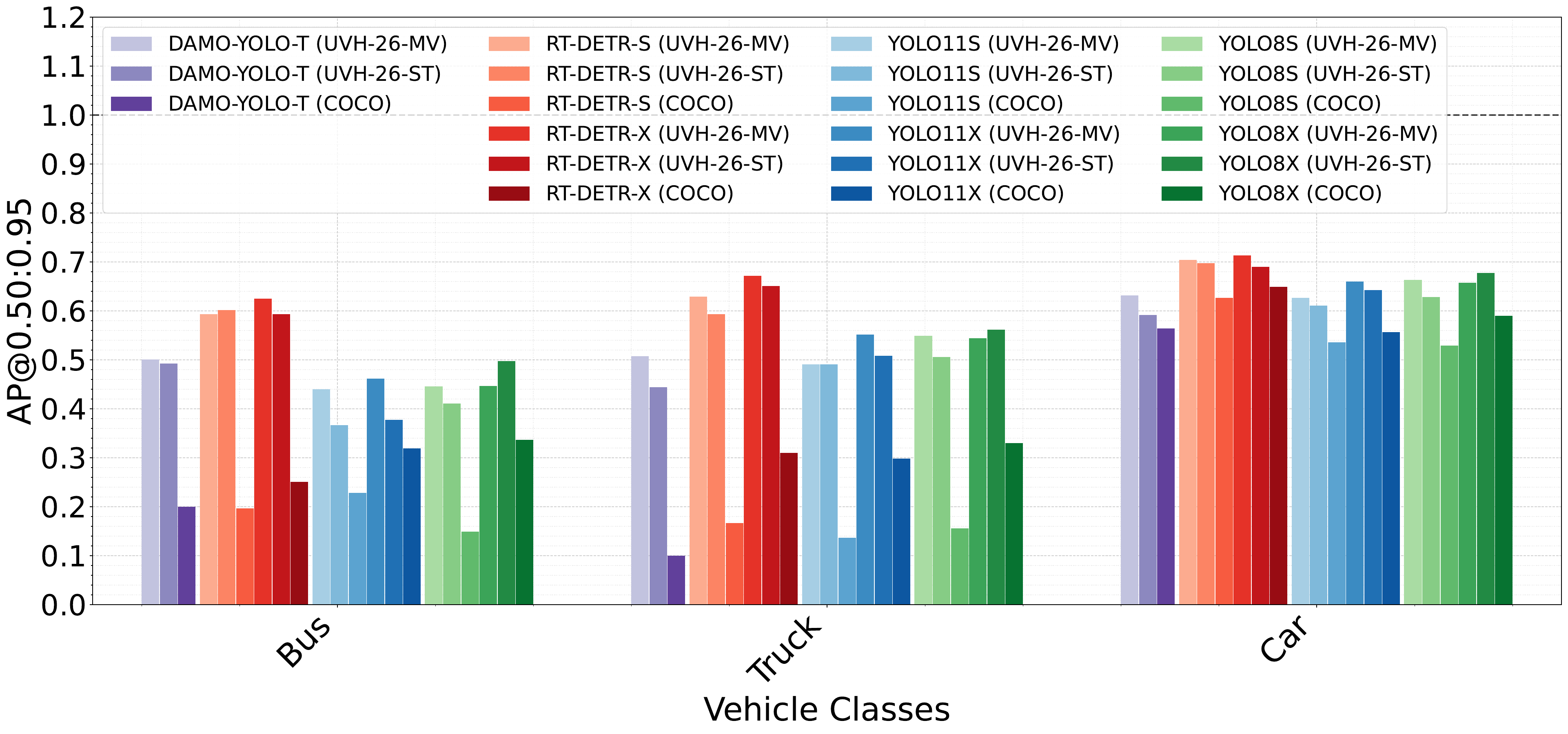}
    \caption{AP@50:95 performance comparison across different models (COCO baseline, \uvhmv, \uvhst) and model families, for the vehicle classes common to COCO and \uvh (Car, Bus, Truck). \uvh models are trained on the \textbf{non-anonymized datasets}.}
    \label{fig:app-ap_50_95_performance_2}
\end{figure}

\begin{figure}[t]
    \centering
    \includegraphics[width=1\textwidth]{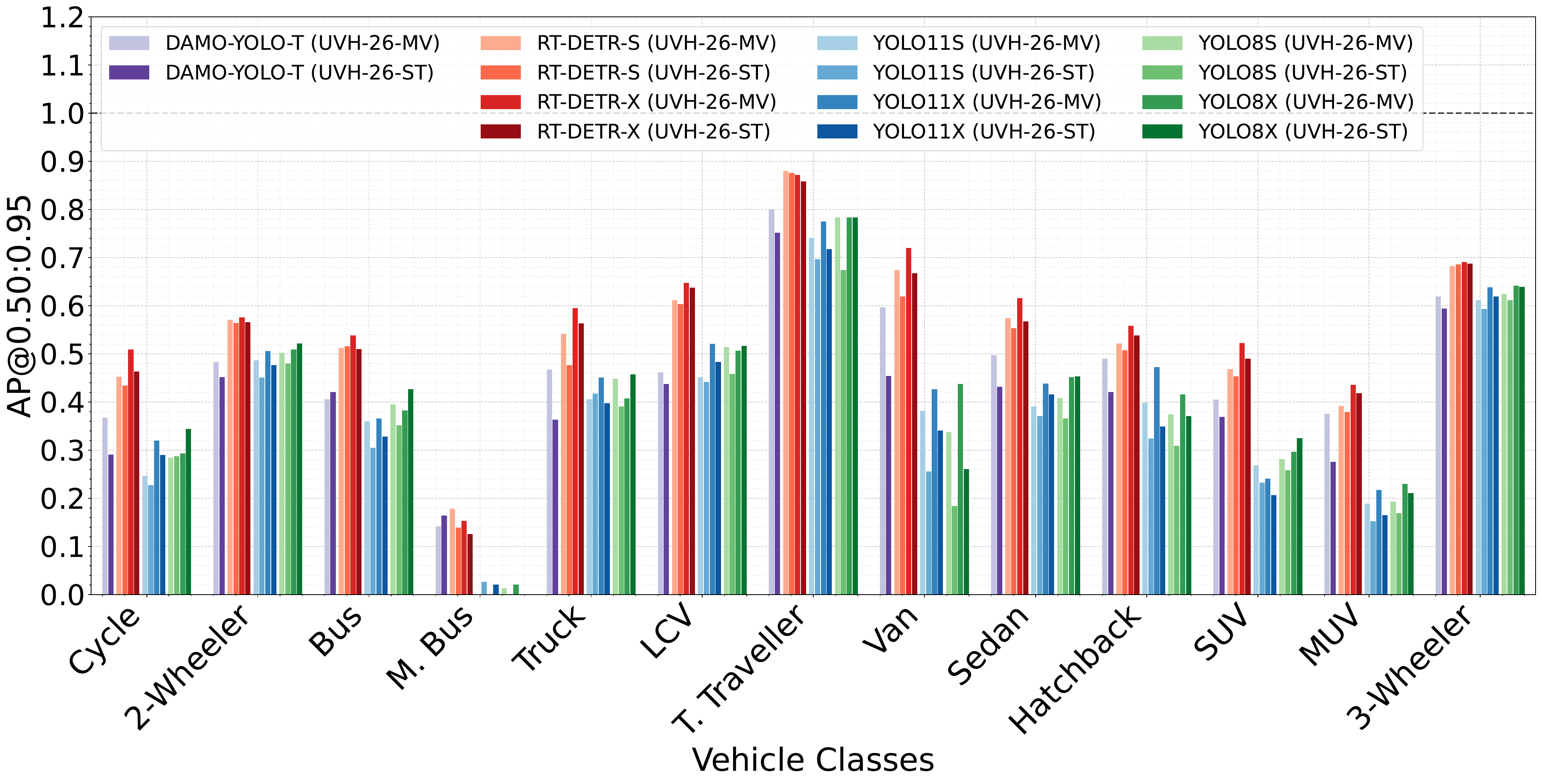}
    \caption{Per-class performance (AP@50:95) of different models and consensus algorithms for the \uvh classes on the \textbf{non-anonymized} \uvhw dataset.}
    \label{fig:app-ap_50_95_performance_3}
\end{figure}

\definecolor{lowval}{HTML}{E74C3C}
\definecolor{midval}{HTML}{F1C40F}
\definecolor{highval}{HTML}{2ECC71}

\newcommand{\valcolor}[1]{%
  \ifdim#1 pt < 0.31 pt
    \textcolor{lowval}{#1}%
  \else
    \ifdim#1 pt < 0.61 pt
      \textcolor{midval}{#1}%
    \else
      \textcolor{highval}{#1}%
    \fi
  \fi
}

\begin{sidewaystable}[htbp]
\centering
\caption{Per-class AP for all \uvh classes for different models trained on the \textbf{non-anonymized} \uvhmv and \uvhst datasets~\tablefootnote{Green color is best score and blue is second best score for a class and AP (maxima in a row), for each AP metric.}.}
\resizebox{\textwidth}{!}{
\begin{tabular}{llcccccccccccccccccccccc}
\toprule
& & \multicolumn{3}{c}{YOLO8S} & \multicolumn{3}{c}{YOLO8X} & \multicolumn{3}{c}{YOLO11S} & \multicolumn{3}{c}{YOLO11X} & \multicolumn{3}{c}{RT-DETR-S} & \multicolumn{3}{c}{RT-DETR-X} & \multicolumn{3}{c}{DAMO-YOLO-T} \\
\cmidrule(lr){3-5} \cmidrule(lr){6-8} \cmidrule(lr){9-11} \cmidrule(lr){12-14} \cmidrule(lr){15-17} \cmidrule(lr){18-20} \cmidrule(lr){21-23}
Class & Method & 50 & 75 & 50:95 & 50 & 75 & 50:95 & 50 & 75 & 50:95 & 50 & 75 & 50:95 & 50 & 75 & 50:95 & 50 & 75 & 50:95 & 50 & 75 & 50:95 \\
\midrule

\multirow{2}{*}{Cycle} & Majority Voting &
0.38 & 0.33 & 0.28 &
0.41 & 0.34 & 0.29 &
0.33 & 0.30 & 0.25 &
0.44 & 0.39 & 0.32 &
0.71 & 0.50 & 0.45 &
\cellcolor{best_color}\textbf{0.79} & \cellcolor{best_color}\textbf{0.56} & \cellcolor{best_color}\textbf{0.51} &
0.59 & 0.41 & 0.37 \\
& STAPLE &
0.40 & 0.34 & 0.29 &
0.48 & 0.39 & 0.34 &
0.31 & 0.27 & 0.23 &
0.42 & 0.34 & 0.29 &
0.68 & 0.47 & 0.43 &
\cellcolor{best2_color}\textbf{0.70} & \cellcolor{best2_color}\textbf{0.49} & \cellcolor{best2_color}\textbf{0.46} &
0.49 & 0.32 & 0.29 \\
\addlinespace[0.6em]

\multirow{2}{*}{2-Wheeler} & Majority Voting &
0.79 & 0.56 & 0.50 &
0.79 & 0.56 & 0.51 &
0.77 & 0.53 & 0.49 &
0.78 & 0.57 & 0.51 &
0.89 & 0.63 & 0.57 &
\cellcolor{best_color}\textbf{0.90} & \cellcolor{best_color}\textbf{0.64} & \cellcolor{best_color}\textbf{0.58} &
0.83 & 0.50 & 0.48 \\
& STAPLE &
0.76 & 0.53 & 0.48 &
0.81 & 0.59 & 0.52 &
0.72 & 0.50 & 0.45 &
0.77 & 0.52 & 0.48 &
\cellcolor{best2_color}\textbf{0.89} & 0.62 & 0.56 &
\cellcolor{best2_color}\textbf{0.89} & \cellcolor{best2_color}\textbf{0.63} & \cellcolor{best2_color}\textbf{0.57} &
0.80 & 0.46 & 0.45 \\
\midrule

\multirow{2}{*}{Bus} & Majority Voting &
0.50 & 0.46 & 0.40 &
0.49 & 0.44 & 0.38 &
0.47 & 0.43 & 0.36 &
0.47 & 0.42 & 0.37 &
0.68 & 0.55 & 0.51 &
\cellcolor{best_color}\textbf{0.71} & \cellcolor{best_color}\textbf{0.58} & \cellcolor{best_color}\textbf{0.54} &
0.58 & 0.44 & 0.41 \\
& STAPLE &
0.44 & 0.41 & 0.35 &
0.52 & 0.48 & 0.43 &
0.39 & 0.33 & 0.30 &
0.42 & 0.36 & 0.33 &
0.67 & \cellcolor{best_color}\textbf{0.58} & \cellcolor{best2_color}\textbf{0.52} &
\cellcolor{best2_color}\textbf{0.69} & 0.56 & 0.51 &
0.62 & 0.45 & 0.42 \\
\addlinespace[0.6em]

\multirow{2}{*}{M. Bus} & Majority Voting &
0.01 & 0.01 & 0.01 &
0.03 & 0.03 & 0.02 &
0.00 & 0.00 & 0.00 &
0.00 & 0.00 & 0.00 &
\cellcolor{best_color}\textbf{0.24} & \cellcolor{best_color}\textbf{0.23} & \cellcolor{best_color}\textbf{0.18} &
0.20 & 0.18 & 0.15 &
0.21 & 0.18 & 0.14 \\
& STAPLE &
0.00 & 0.00 & 0.00 &
0.00 & 0.00 & 0.00 &
0.03 & 0.03 & 0.03 &
0.03 & 0.03 & 0.02 &
0.17 & 0.17 & 0.14 &
0.16 & 0.16 & 0.13 &
\cellcolor{best2_color}\textbf{0.22} & \cellcolor{best2_color}\textbf{0.20} & \cellcolor{best2_color}\textbf{0.16} \\
\midrule

\multirow{2}{*}{Truck} & Majority Voting &
0.58 & 0.55 & 0.45 &
0.54 & 0.51 & 0.41 &
0.56 & 0.50 & 0.41 &
0.59 & 0.55 & 0.45 &
0.72 & 0.66 & 0.54 &
\cellcolor{best_color}\textbf{0.77} & \cellcolor{best_color}\textbf{0.72} & \cellcolor{best_color}\textbf{0.59} &
0.70 & 0.55 & 0.47 \\
& STAPLE &
0.53 & 0.47 & 0.39 &
0.61 & 0.52 & 0.46 &
0.54 & 0.51 & 0.42 &
0.54 & 0.51 & 0.40 &
0.63 & 0.58 & 0.48 &
\cellcolor{best2_color}\textbf{0.73} & \cellcolor{best2_color}\textbf{0.66} & \cellcolor{best2_color}\textbf{0.56} &
0.59 & 0.38 & 0.36 \\
\addlinespace[0.6em]

\multirow{2}{*}{LCV} & Majority Voting &
0.69 & 0.59 & 0.51 &
0.69 & 0.59 & 0.51 &
0.62 & 0.52 & 0.45 &
0.69 & 0.60 & 0.52 &
0.84 & 0.74 & 0.61 &
\cellcolor{best_color}\textbf{0.87} & \cellcolor{best_color}\textbf{0.78} & \cellcolor{best_color}\textbf{0.65} &
0.74 & 0.50 & 0.46 \\
& STAPLE &
0.62 & 0.53 & 0.46 &
0.69 & 0.60 & 0.52 &
0.62 & 0.51 & 0.44 &
0.69 & 0.55 & 0.48 &
0.81 & 0.71 & 0.60 &
\cellcolor{best2_color}\textbf{0.85} & \cellcolor{best2_color}\textbf{0.76} & \cellcolor{best2_color}\textbf{0.64} &
0.71 & 0.47 & 0.44 \\
\midrule

\multirow{2}{*}{T. Traveller} & Majority Voting &
0.83 & 0.83 & 0.78 &
0.83 & 0.83 & 0.78 &
0.81 & 0.81 & 0.74 &
0.82 & 0.81 & 0.77 &
\cellcolor{best_color}\textbf{0.96} & \cellcolor{best_color}\textbf{0.92} & \cellcolor{best_color}\textbf{0.88} &
\cellcolor{best_color}\textbf{0.96} & 0.91 & 0.87 &
0.90 & 0.89 & 0.80 \\
& STAPLE &
0.72 & 0.72 & 0.67 &
0.83 & 0.83 & 0.78 &
0.75 & 0.74 & 0.70 &
0.77 & 0.77 & 0.72 &
\cellcolor{best2_color}\textbf{0.95} & \cellcolor{best_color}\textbf{0.92} & \cellcolor{best_color}\textbf{0.88} &
0.94 & 0.90 & 0.86 &
0.86 & 0.84 & 0.75 \\
\addlinespace[0.6em]

\multirow{2}{*}{Van} & Majority Voting &
0.40 & 0.40 & 0.34 &
0.52 & 0.51 & 0.44 &
0.45 & 0.45 & 0.38 &
0.51 & 0.51 & 0.43 &
0.83 & 0.81 & 0.67 &
\cellcolor{best_color}\textbf{0.88} & \cellcolor{best_color}\textbf{0.85} & \cellcolor{best_color}\textbf{0.72} &
0.78 & 0.75 & 0.60 \\
& STAPLE &
0.21 & 0.21 & 0.18 &
0.30 & 0.30 & 0.26 &
0.31 & 0.31 & 0.26 &
0.41 & 0.41 & 0.34 &
0.75 & 0.74 & 0.62 &
\cellcolor{best2_color}\textbf{0.81} & \cellcolor{best2_color}\textbf{0.80} & \cellcolor{best2_color}\textbf{0.67} &
0.60 & 0.57 & 0.45 \\
\midrule

\multirow{2}{*}{Sedan} & Majority Voting &
0.50 & 0.48 & 0.41 &
0.56 & 0.52 & 0.45 &
0.49 & 0.45 & 0.39 &
0.54 & 0.51 & 0.44 &
0.74 & 0.67 & 0.57 &
\cellcolor{best_color}\textbf{0.78} & \cellcolor{best_color}\textbf{0.71} & \cellcolor{best_color}\textbf{0.62} &
0.68 & 0.58 & 0.50 \\
& STAPLE &
0.45 & 0.42 & 0.37 &
0.55 & 0.53 & 0.45 &
0.45 & 0.43 & 0.37 &
0.53 & 0.49 & 0.42 &
0.70 & 0.65 & 0.55 &
\cellcolor{best2_color}\textbf{0.72} & \cellcolor{best2_color}\textbf{0.66} & \cellcolor{best2_color}\textbf{0.57} &
0.61 & 0.50 & 0.43 \\
\addlinespace[0.6em]

\multirow{2}{*}{Hatchback} & Majority Voting &
0.45 & 0.44 & 0.37 &
0.50 & 0.48 & 0.42 &
0.48 & 0.47 & 0.40 &
0.57 & 0.55 & 0.47 &
0.63 & 0.61 & 0.52 &
\cellcolor{best_color}\textbf{0.68} & \cellcolor{best_color}\textbf{0.65} & \cellcolor{best_color}\textbf{0.56} &
0.63 & 0.59 & 0.49 \\
& STAPLE &
0.37 & 0.36 & 0.31 &
0.45 & 0.43 & 0.37 &
0.39 & 0.38 & 0.32 &
0.43 & 0.41 & 0.35 &
0.61 & 0.59 & 0.51 &
\cellcolor{best2_color}\textbf{0.66} & \cellcolor{best2_color}\textbf{0.63} & \cellcolor{best2_color}\textbf{0.54} &
0.55 & 0.51 & 0.42 \\
\addlinespace[0.6em]

\multirow{2}{*}{SUV} & Majority Voting &
0.33 & 0.32 & 0.28 &
0.35 & 0.33 & 0.30 &
0.32 & 0.31 & 0.27 &
0.28 & 0.27 & 0.24 &
0.57 & 0.53 & 0.47 &
\cellcolor{best_color}\textbf{0.64} & \cellcolor{best_color}\textbf{0.60} & \cellcolor{best_color}\textbf{0.52} &
0.53 & 0.46 & 0.41 \\
& STAPLE &
0.30 & 0.29 & 0.26 &
0.38 & 0.37 & 0.32 &
0.27 & 0.26 & 0.23 &
0.24 & 0.23 & 0.21 &
0.54 & 0.51 & 0.45 &
\cellcolor{best2_color}\textbf{0.59} & \cellcolor{best2_color}\textbf{0.55} & \cellcolor{best2_color}\textbf{0.49} &
0.47 & 0.43 & 0.37 \\
\addlinespace[0.6em]

\multirow{2}{*}{MUV} & Majority Voting &
0.24 & 0.23 & 0.19 &
0.29 & 0.26 & 0.23 &
0.23 & 0.22 & 0.19 &
0.26 & 0.25 & 0.22 &
0.49 & 0.47 & 0.39 &
\cellcolor{best_color}\textbf{0.55} & \cellcolor{best_color}\textbf{0.52} & \cellcolor{best_color}\textbf{0.44} &
0.50 & 0.44 & 0.38 \\
& STAPLE &
0.21 & 0.19 & 0.17 &
0.26 & 0.24 & 0.21 &
0.19 & 0.17 & 0.15 &
0.20 & 0.18 & 0.17 &
0.47 & 0.45 & 0.38 &
\cellcolor{best2_color}\textbf{0.52} & \cellcolor{best2_color}\textbf{0.50} & \cellcolor{best2_color}\textbf{0.42} &
0.39 & 0.31 & 0.28 \\
\midrule

\multirow{2}{*}{3-Wheeler} & Majority Voting &
0.86 & 0.71 & 0.62 &
0.88 & 0.74 & 0.64 &
0.85 & 0.70 & 0.61 &
0.88 & 0.74 & 0.64 &
0.93 & 0.79 & 0.68 &
\cellcolor{best_color}\textbf{0.94} & \cellcolor{best_color}\textbf{0.80} & \cellcolor{best_color}\textbf{0.69} &
0.91 & 0.71 & 0.62 \\
& STAPLE &
0.84 & 0.70 & 0.61 &
0.88 & 0.74 & 0.64 &
0.82 & 0.68 & 0.59 &
0.87 & 0.70 & 0.62 &
0.93 & 0.79 & \cellcolor{best_color}\textbf{0.69} &
\cellcolor{best_color}\textbf{0.94} & \cellcolor{best_color}\textbf{0.80} & \cellcolor{best_color}\textbf{0.69} &
0.90 & 0.67 & 0.59 \\

\bottomrule
\end{tabular}
}
\label{tab:hackathon_per_class_merged}
\end{sidewaystable}

\begin{table}[ht]
    \centering
    \caption{\textit{Overall mAP metrics} across all \uvh classes per model. Each mAP value is averaged over the per-class AP for all \uvh classes~\tablefootnote{Green color is best score and blue is second best score for column.}.}
    \footnotesize
    \begin{tabular}{lcrrr}
        \toprule
        \bf Model & \bf Method & \bf mAP(50) & \bf mAP(50:95) & \bf mAP(75) \\
        \hline\hline

        YOLO8S     & Majority Voting & 0.51 & 0.40 & 0.45 \\
        YOLO8X     & Majority Voting & 0.53 & 0.41 & 0.47 \\
        YOLO11S    & Majority Voting & 0.49 & 0.38 & 0.44 \\
        YOLO11X    & Majority Voting & 0.53 & 0.41 & 0.47 \\
        RT-DETR-S  & Majority Voting & 0.71 & 0.54 & 0.62 \\
        RT-DETR-X  & Majority Voting & \cellcolor{best_color}\textbf{0.74} & \cellcolor{best_color}\textbf{0.57} & \cellcolor{best_color}\textbf{0.65} \\
        DAMO-YOLO-T & Majority Voting & 0.66 & 0.47 & 0.54 \\

        \midrule
         \addlinespace[0.8em]

        YOLO8S     & STAPLE & 0.45 & 0.35 & 0.40 \\
        YOLO8X     & STAPLE & 0.52 & 0.41 & 0.46 \\
        YOLO11S    & STAPLE & 0.45 & 0.35 & 0.39 \\
        YOLO11X    & STAPLE & 0.49 & 0.37 & 0.42 \\
        RT-DETR-S  & STAPLE & 0.68 & 0.52 & 0.60 \\
        RT-DETR-X  & STAPLE & \cellcolor{best2_color}\textbf{0.71} & \cellcolor{best2_color}\textbf{0.55} & \cellcolor{best2_color}\textbf{0.62} \\
        DAMO-YOLO-T & STAPLE & 0.60 & 0.42 & 0.47 \\

        \bottomrule
    \end{tabular}
    \label{tab:hackathon_overall_map}
\end{table}

\end{document}